 \documentclass[final,5p,times,twocolumn]{elsarticle}

\usepackage{amssymb}
\usepackage{booktabs}
\usepackage{xcolor}
\usepackage{color}
\usepackage{textcomp}
\usepackage{amssymb}
\usepackage{algorithm}
\usepackage{algpseudocode}
\usepackage{stmaryrd}
\usepackage{stackrel}
\usepackage{multirow}
\usepackage{titlesec}
\usepackage{enumitem}
\usepackage{mathtools}
\usepackage{fancyhdr}
\usepackage{caption}
\usepackage{amsthm}
\usepackage{hyperref} 
\usepackage{subcaption}

\journal{.}

\begin{document}

\begin{frontmatter}



\title{Remote sensing data imputation using deep learning for multispectral imagery}


\author[inst1,inst2]{Shuang Liu}

\affiliation[inst1]{organization={Water Research Centre, University of New South Wales},
      city={Sydney}, 
      state={NSW},
      postcode={2052},
      country={Australia}}

\affiliation[inst2]{organization={ARC ITTC Data Analytics for Resources and Environments, University of New South Wales},
      city={Sydney}, 
      state={NSW},
      postcode={2052},
      country={Australia}}

\author[inst1,inst2]{Fiona Johnson}
\author[inst2,inst3]{Rohitash Chandra}

\affiliation[inst3]{organization={Transitional Artificial Intelligence Research Group, School of Mathematics and Statistics, University of New South Wales},
      city={Sydney}, 
      state={NSW},
      postcode={2052},
      country={Australia}}

\begin{abstract}

Remote sensing techniques have been increasingly utilised in aquatic applications in recent years. A common challenge in using optical satellite data is the presence of missing observations due to cloud cover. These data gaps can lead to missed detection of critical events, such as algal blooms, in lakes of high interest to water authorities. As a result, enhancing the completeness of optical satellite datasets is crucial for improving the monitoring and prediction of algal blooms. In this study, we compared a traditional data imputation method (i.e., linear interpolation) with deep learning models for reconstructing missing spectral bands across four lakes with historical records of algal blooms. The deep learning models adopted include CNN-based architectures (i.e., CNN, Inception Resnet, and Autoencoder) and CNN-LSTM-based architectures (i.e., CNN-LSTM, Resnet-LSTM, and Autoencoder-LSTM). Our results demonstrated that deep learning models substantially outperformed the baseline linear interpolation method in imputing spectral band values within artificially masked regions. Among these models, CNN delivered the best performance across most lakes. Furthermore, we evaluated the performance of algal bloom indices (i.e., Green/Red and NDCI) derived from the imputed imagery by comparing them with the observed data. Our results demonstrate that deep learning models are effective for imputing missing data in PlanetScope SuperDove imagery, enabling more reliable applications in water monitoring. 

\end{abstract}



\end{frontmatter}



\section{Introduction}

Remote sensing has become an increasingly valuable tool for aquatic monitoring in recent decades \cite{matthews2017bio,houborg2018cubesat, verpoorter2014global,chawla2020review} supporting assessments of water quantity and quality in estuaries \cite{le2013evaluation}, large lakes \cite{ho2019widespread}, and coastal waters \cite{gholizadeh2016comprehensive}. Satellite-based observations provide broad spatial coverage and relatively high temporal resolution, making them especially useful in regions where field data is sparse or unavailable \cite{liu2022remote}. The emergence of CubeSat satellites \cite{marta2018planet} such as those operated by Planet, has further enhanced monitoring capabilities by providing improved spatial, temporal and spectral resolution. These advancements offer new opportunities for lake management, including tracking changes in water quality and quantity \cite{bareuther2020spatio, beck2017comparison}. As a result, there is a growing demand for satellite-based spatiotemporal monitoring to support informed decision-making in lake and catchment management.  

Despite the growing utility of satellite observations, their effectiveness is often limited by factors that compromise data quality and lead to gaps in data coverage \cite{daniels2022filling}. These include technical issues, such as instrument malfunctions (e.g., scan line corrector failure in Landsat 7, and missed detection in GRACE due to battery issues)  \cite{scaramuzza2005landsat,yi2021filling}, adverse weather conditions (i.e., cloud cover) \cite{liu2022remote}, and satellite orbit constraints that result in missing swaths \cite{frazier2021technical}. In optical satellite, the missing values caused by natural events such as cloud cover can hinder monitoring, prediction and decision-making. This is especially critical in aquatic applications, where data gaps obscure key areas of lakes undergoing algal blooms. Given the high spatial variability of blooms \cite{liu2022remote}, even partial cloud cover can prevent accurate detection and assessment, underscoring the need for effective data imputation. 

PlanetScope \footnote{\url{https://earth.esa.int/eogateway/missions/planetscope}} (launched in 2018) has been utilised in various earth observation applications, such as land-cover mapping  \cite{vizzari2022planetscope}, crop yield estimation \cite{sagan2021field}, and monitoring water quantity and quality  \cite{ehret2021automatic, mansaray2021comparing}. The latest addition to PlanetScope, the SuperDove (PSB.SD), has been in operation since 2020  \cite{Planet2023TechReport}. PSB.SD imagery offers high spatial resolution (3 meters), frequent temporal coverage (sub-daily -- daily) and expanded spectral bands (8 bands), including visible, and near-infrared wavelengths, along with additional bands (i.e., yellow and Green 1). Its spectral calibration has been refined to closely match Sentinel 2 \cite{tu2022radiometric}, facilitating the transferability of algorithms and methods originally developed for Sentinel 2 to PBS.SD. However, the presence of cloud-masked missing pixels remains a major challenge, limiting the effectiveness of PlanetScope SuperDove. To maximise the utility of optical satellite imagery and improve model performance, pixel-level data imputation is essential, particularly for capturing dynamic changes in lake systems affected by deteriorating water quality, such as turbidity and algal blooms. 

The application of advanced deep learning for imputing missing values in the emerging PBS.SD imagery remains limited in the literature. Different approaches have been proposed for gap-filling in spatiotemporal datasets, encompassing statistical methods and deep learning methods \cite{shen2015missing}. Statistical methods include discrete cosine transforms \cite{wang2012three}, singular spectrum analysis \cite{yi2021filling}, spatiotemporal interpolation \cite{appel2020spatiotemporal}, and algorithmic methods (e.g., quantile regression in local neighbourhoods) \cite{gerber2018predicting}. Deep learning methods have gained significant traction in remote sensing, with the increasing availability of hyperspectral and multispectral satellite imagery \cite{paheding2024advancing, pritt2017satellite, waldner2020deep}. Deep learning models such as convolutional neural networks (CNNs) have been widely employed for satellite imagery analysis tasks, including object detection and segmentation \cite{segal2020cloud, zhang2021rich}, and spatiotemporal modelling  \cite{barzegar2020short, chen2020deep, mo2022bayesian}. CNNs have been effective at processing multi-dimensional images and extracting features with minimal preprocessing, making them superior in image prediction  \cite{archana2024deep}. Deep learning models, such as Autoencoders and U-Net, have been promising in restoring corrupted portions of satellite images \cite{cresson2019optical, daniels2022filling, liang2023reconstructing, liu2018image,appel2024efficient, lops2021application, xing2022spatiotemporal}. Autoencoder-based models learn a representation of the data from the input layer and try to reproduce it at the output layer, and therefore learn from incomplete data and generate new plausible values for imputation. Inception-Resnet combines the multi-scale feature extraction of the Inception architecture with the efficient training by Resnet's residual learning \cite{szegedy2017inception}. Furthermore, hybrid models, such as CNN -- Long Short-Term Memory Network (CNN-LSTM) recurrent neural network architectures, have been successfully applied to fill the cloud-induced gaps in satellite data \cite{daniels2022filling, qian2024gap}. The strength of CNN-LSTM models lies in their ability to capture both spatial and temporal features by integrating CNNs for spatial feature extraction and LSTMs for sequential data modelling. 

Gap-filling and image reconstruction have been extensively applied to multispectral images from MODIS (Moderate Resolution Imaging Spectroradiometer)  \cite{moreno2020multispectral, roy2008multi, wang2022new}, Landsat \cite{chen2011simple, yin2016gap}, and Sentinel-2 \cite{cresson2019optical} satellite data. However, studies on multispectral gap filling for PlanetScope SuperDove are limited  \cite{wang2022new}. Data imputation is further complicated by the inherent complexity of remote sensing data formats, which often include multiple spectral bands (e.g., multi-spectral and hyper-spectral data). Effective data imputation methods need to account for both spatial dependencies and spectral variability to accurately reconstruct missing data, generating plausible representations of the same object across different spectral dimensions. 

In this study, we present a machine learning data imputation framework that involves efficient and high-performance deep learning models to impute missing regions in remote sensing data. Our objective is to evaluate the model performance of data imputation for PlanetScope SuperDove imagery affected by cloud masks across different lake types. To achieve this, we first developed and examined selected deep learning models for data imputation of PlanetScope SuperDove imagery with artificial cloud masks across all eight spectral bands for lakes. We evaluated prominent deep learning models, including CNN, hybrid LSTM-CNN, ResNet, and hybrid Autoencoder-LSTM, in our machine learning framework. Subsequently, we applied the imputed PlanetScope SuperDove imagery to derive water quality indices by comparing a baseline model with the deep learning models across four Australian lakes with diverse characteristics in size and water quality conditions: Lake Burragorang, Lake Hume, Lake Grahamstown and Lake Carramar.

\section{Data and Methodology}

 \subsection{Data}
 \subsubsection{Site description}

Four lakes, ranging from 7274 m\textsuperscript{2} to 200 km\textsuperscript{2} were selected to assess the effectiveness of deep learning models for data imputation across different lake types. These lakes differ in the frequency of harmful algal bloom occurrences \cite{king2022murray, liu2022remote, liu2021effectiveness, vilhena2010role}. 

\begin{figure*} [htbp!]
\centering
\includegraphics[width=18cm]{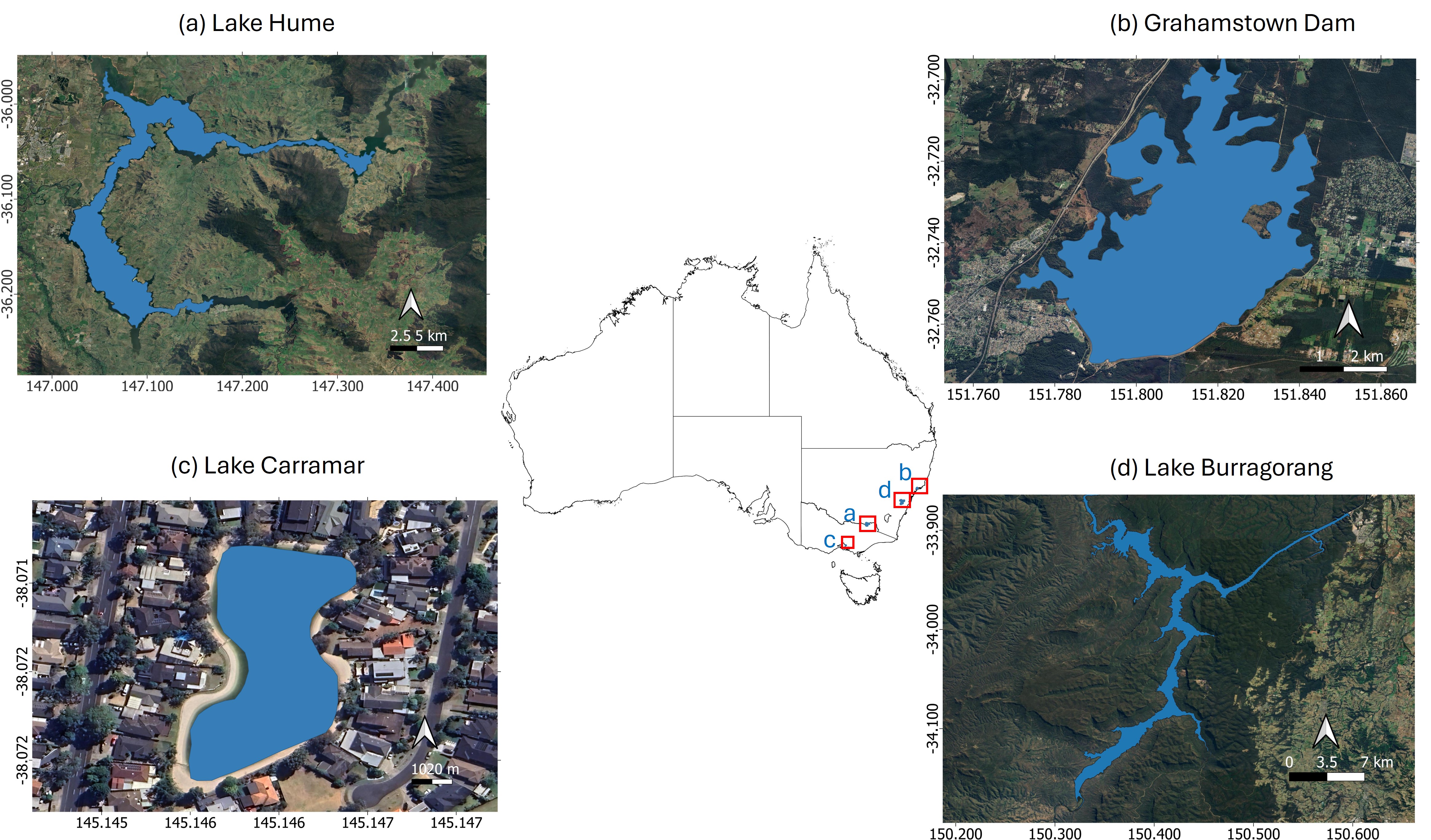}
\caption{ Location of a) Lake Hume, b) Lake Grahamstown, c) Lake Carramar, and d) Lake Burragorang, Australia}
\label{fig:location}
\end{figure*}

Lake Hume (Figure~\ref{fig:location}a) is a major reservoir on the Murray River, located in the Murray-Darling Basin, Australia. As the major operational storage for the Murray River, it supplies water for irrigation, domestic, stock and urban demands to Victoria and New South Wales. Covering an area of 200 km\textsuperscript{2}, the lake has a maximum water depth of 40 metres. This lake has been frequently reported to experience algal blooms \cite{king2022murray}. 

Lake Grahamstown (Figure~\ref{fig:location}b) is a mesotrophic freshwater lake that has experienced periodic cyanobacterial blooms \cite{golshan2020patterns,mueller2014role}. It is the largest \href{https://www.sciencedirect.com/topics/earth-and-planetary-sciences/potable-water}{drinking water} storage in the lower Hunter River valley \cite{water2011catchment}. The lake receives an average annual rainfall of 1125 mm and is relatively shallow, averaging 9 metres, with a surface area of 28 km\textsuperscript{2}. The dam is also supplied by diversions from the nearby Williams River, which has highly variable water quality. 

Lake Carramar (Figure~\ref{fig:location}c) is a small artificial lake, located in Melbourne. Surrounded by a residential area, Lake Carramar is interlinked with two non-tidal lakes, forming the Quiet Lakes system. Lake Carramar is a key part of the local drainage system, receiving stormwater that drains into the nearby Patterson River. Lake Carramar covers an area of 7274 m\textsuperscript{2}, with an average depth of 3m. Lake Carramar serves as a stormwater and recreational lake and frequently experiences turbidity and harmful algal bloom events.

Lake Burragorong (Figure~\ref{fig:location}d) is a reservoir situated in the lower Blue Mountains of New South Wales, Australia. Covering an area of 75 km\textsuperscript{2}, it is the largest water supply for Greater Sydney, providing nearly 80\% of the water supply of the region's water  \cite{WaterNSW2015TechReport}. The maximum depth of the lake is 105m. 

\begin{table*}
\centering
\small
\caption{Details of lake size, average depth, function and harmful algal bloom occurrence in four lakes and period (days) considered. }
\label{tab:lake}
\begin{tabular}{| l | l | l | l | l | l |}
\hline
 & Lake size & Average depth & Function & Harmful Algal bloom & Period (days) \\
\hline
Lake Carramar & 7274 m\textsuperscript{2} & 3 m & Stormwater & Frequent & Sep 2020-Dec 2024 (215)\\
\hline
Lake Grahamstown & 28 km\textsuperscript{2} & 9 m & Dam & Less frequent & Sep 2020-Dec 2024 (335)\\
\hline
Lake Burragorang & 75 km\textsuperscript{2} & 9 m & Dam & Less frequent & Aug 2020 - Dec 2024 (240)\\
\hline
Lake Hume & 200 km\textsuperscript{2} & 27 m & Recreational & Frequent & Aug 2020 - Dec 2024 (225)\\
\hline

\end{tabular}

\end{table*}

\subsubsection{Remote sensing data}

PlanetScope SuperDove (PSB.SD) is a new generation CubeSat satellite constellation with very high spatial (\~3m) and frequent temporal coverage (daily or sub-daily)  \cite{Planet2023TechReport}, providing new opportunities for monitoring water quality in lakes and reservoirs. Available since June 2020, PSB.SD features eight spectral bands, including the distinctive Green 1 and Yellow bands, compared with other satellites (i.e., Sentinel 2). Planetscope SuperDove imagery was acquired through Planet's API and downloaded as orthorectified, harmonised surface reflectance images in GeoTIFF format.  

\begin{table*}
\centering
\small
\caption{Summary of satellite products used for analysis.}
\label{tab:satellite}
\begin{tabular}{| l |l |l |}
\hline
 & \multicolumn{2}{|l|}{Planetscope SuperDove (PBS.SD)} \\
\hline
Resolution (m) & \multicolumn{2}{|l|}{3} \\
\hline
Revisit time& \multicolumn{2}{|l|}{Sub-daily to daily} \\
\hline
Data availability & \multicolumn{2}{|l|}{August 2020 - present} \\
\hline
Spectral bands& & Central wavelength (nm) \\
\hline
 & Blue & 490 (50) \\ \hline 
 & Green I & 531 (36) \\ \hline 
 & Green & 565 (36) \\ \hline 
 & Yellow & 610 (20) \\ \hline 
 & Red & 665 (31) \\ \hline 
 & Red Edge & 705 (15) \\ \hline 
 & NIR & 865 (40) \\ \hline 
\end{tabular}

\end{table*}

 \subsection{Methods}

\subsection{Deep learning models}

We employed two primary variants of deep learning models, including 1) CNN-based models and 2) CNN-LSTM-based models. In particular, our models include CNN, Inception ResNet CNN, Autoencoder CNN, CNN-LSTM, Inception ResNet LSTM, and Autoencoder LSTM (Figure~\ref{fig:framework}). Additionally, we applied a simple linear interpolation method as a baseline for comparison.

\subsubsection{CNN-based models}

CNNs consist of multiple convolutional and pooling layers designed to efficiently extract spatial features \cite{alzubaidi2021review} and learn patterns through parameterised convolutional filters. CNNs have been prominent for processing image and video data and remote sensing applications \cite{han2023survey,shirmard2022review}. In this study, we design a CNN for spatial feature extraction from PSB.SD imagery, with a custom loss function that selectively calculates MSE (Mean Squared Error) for artificially masked pixel values (see Equation 1). The CNN model architecture features multiple convolutional and pooling layers designed for effective spatial feature extraction across multiple spectral bands, with further details shown in Table 3.

An autoencoder is a neural network with a symmetrical structure consisting of encoding and decoding layers \cite{li2023comprehensive}, where the number of input and output variables is equal. It is designed to learn an efficient data representation (encoding) for dimensionality reduction. In this study, we developed an autoencoder-based CNN (Autoencoder-CNN) for PSB.SD imputation. 

The model architecture includes three main components: an encoder, a bottleneck layer and a decoder. The encoder processed both the input \textbf{X} and the artificial mask, using convolutional layers followed by max-pooling layers. The number of filters in the convolutional layers increases progressively with depth, starting with an initial filter count and doubling at each additional layer. The bottleneck layer comprises a convolutional layer with a higher filter count, capturing the essential features of X, and creating a compressed representation. The decoder mirrors the encoder structure. Each decoder layer uses upsampling to scale feature maps back to the original size. 

The Inception ResNet model architecture \cite{szegedy2017inception}, originally pre-trained on large datasets, is designed to capture robust hierarchical features, making it suitable for satellite image reconstruction. ResNet applied a residual learning strategy to increase the depth of its network. The Inception Resnet showed powerful performance in satellite image classification \cite{alotaibi2020hybrid, iyer2021deep}. We employed Inception ResNet to create a computationally efficient CNN capable of processing the full spectral-spatial input from this 8-band PSB.SD imagery. The model differs from conventional RGB images by modifying the model's input layer to handle 8-band data. We used the base Inception-Resnet model without pre-trained weights. On top of the base, we included global average pooling, followed by dropout for regularisation and a dense layer. We used a masked MSE loss function to calculate the MSE loss in the masked pixels. 

\subsubsection{LSTM-based models}

CNN-LSTM \cite{donahue2015long, vinyals2015show} is a hybrid neural network that combines convolutional neural network (CNN) layers with long short-term memory (LSTM) layers in a unified framework. This model has been prominent in time series prediction tasks and has also been applied to remote sensing data processing \cite{boulila2021novel, pan2023cnn}. The paired CNN with LSTM was applied to leverage both spatial and temporal information inherent in remote sensing imagery. Planetscope SuperDove captures repeated observations of the same location over time, forming a timeseries of images. These temporal sequences are particularly valuable for monitoring dynamic water quality events, such as algal blooms, which can evolve rapidly over days. The model was designed to impute missing pixel values across eight bands over time, using a masked MSE loss function that specifically targets the masked (missing) pixels during training. 
The CNN-LSTM is well-suited for this task, as CNN layers extract spatial features from each image (e.g., water quality status, algal bloom shape, and spectral patterns), while LSTM layers model the temporal dependencies across multiple observation dates. Specifically, we used sequences of five consecutive images (timesteps) to train the model. Each timestep was processed independently by convolutional layers with ReLU activation and defined kernel units (see Table 3). After pooling and flattening, the spatial features were passed to the LSTM layer, which captured temporal patterns across the five-image sequence. The model was trained using a masked mean square error (MSE) loss function, which focuses learning on the missing pixels only. The configuration of CNN-LSTM adopted here is detailed in Table 3. 

We also implemented an Autoencoder-LSTM \cite{srivastava2015unsupervised} model, which combines spatial feature extraction and temporal sequence learning within an encoder-decoder framework. Autoencoder-LSTM models have been successfully applied in remote sensing applications \cite{shakeel2022aladdin}. In our approach, the encoder consists of 2D convolutional blocks applied to each timestep using a time-split structure. This ensures that spatial features are extracted from each image in the sequence without temporal interference. Each convolutional layer was followed by a max-pooling layer to reduce the spatial dimensions and highlight key features. The temporal modelling is handled by an LSTM layer, which acts as the bottleneck of the autoencoder. After flattening the spatial features from each timestep, the LSTM captures temporal dependencies across the sequence of five images. To reduce overfitting, a dropout layer was applied after the LSTM. The decoder mirrored the encoder structure, using upsampling layers to restore the spatial dimensions of the input. As the decoder progresses, the number of filters is gradually decreased to match the original input dimensions. A final output layer was applied to reconstruct the imputed image for each timestep, completing the data imputation process. 

Inception-ResNet-LSTM combines the strengths of advanced spatial and temporal extraction, making it well-suited for remote sensing applications involving time series data. Specifically, the model integrates convolutional layers from the Inception ResNet architecture, known for capturing multi-scale spatial features efficiently, with an LSTM layer that models temporal dependencies across sequential images. The Inception Resnet component is wrapped in a time-distributed structure. allowing it to process each image in the time series while maintaining temporal alignment. This is followed by pooling and flattening layers that prepare the extracted spatial features for temporal modelling. The LSTM layer then processed these spatial features across timesteps, learning patterns and changes over time, such as the development of algal blooms. The remote sensing imagery contains complex spatial patterns (e.g., water quality status, algal bloom shapes) and temporal dynamics (e.g., bloom formation, decay and movement). By combining Inception-ResNet's spatial sensitivity with LSTM's temporal modelling, the model can effectively reconstruct missing data and preserve both spatial detail and temporal continuity.

\subsection{Baseline model}

We implemented a simple linear interpolation method as the simple baseline model to compare against other machine learning methods. For each date, across the latitude and longitude for eight channels were calculated and used to fill the missing values in each individual band.

\subsection{Framework for data imputation}

\begin{figure*} [htbp!]
\centering
\includegraphics[width=18 cm]{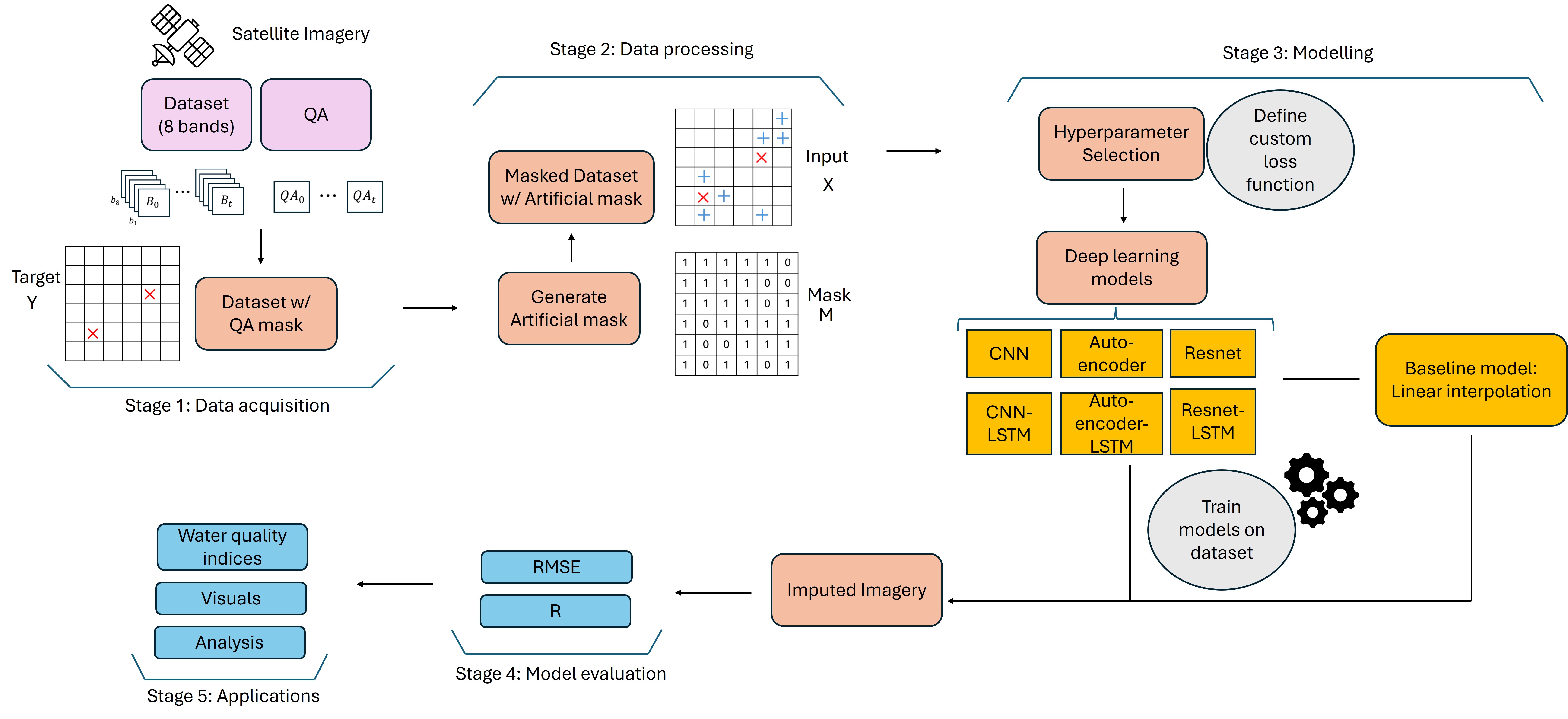}
\caption{Overview of the data imputation and its application in five stages, including data acquisition (8-band imagery and QA data), data processing (artificial mask generation), modelling of deep learning models (i.e., CNN, Autoencoder, Inception Resnet, CNN-LSTM, Autoencoder LSTM, Resnet LSTM) and baseline model (linear interpolation), model evaluation and applications.}
\label{fig:framework}
\end{figure*}

We begin with an overview of our approach for data imputation (Figure~\ref{fig:framework}), followed by detailed explanations for each component in the subsequent sections. The approach consists of five stages: 1) data acquisition of PSB.SD imagery, 2) data processing of PSB.SD data by manually applying data masks to represent clouds and preparing training and test datasets, 3) developing deep learning models and a baseline model for data imputation within the artificial masks, 4) model evaluation of the imputed imagery, and 5) water quality applications of the imputed imagery.

In Stage 1, we acquired eight-band SD surface reflectance data and their QA (Quality Assessment) from the PlanetScope platform. The QA data were used to identify and filter out missing pixels, creating the target dataset \textbf{Y}. We processed the raw SD imagery by applying the Usable Data Mask (UDM) to generate an optimal dataset before training the model for data imputation. The UDM file contains information on weather conditions, including clear, snow, shadow, light haze and cloud  \cite{Planet2023TechReport}. We filter the imagery using the clear masks from UDM files to remove pixels affected by cloud interference. We selected the images with fewer clouds in the dataset, and the number of images for each individual lake is shown in Table 2.

In Stage 2, we generated artificial masks on the selected imagery because the model performance can only be evaluated at the pixels with available data. \textbf{Y} represents the target dataset, which may already contain missing values. \textbf{M\textsubscript{x}} is an artificial binary mask, where 1s denote available pixels and 0s indicate missing values. The methods described in \cite{appel2024efficient} were taken to generate additional gaps in\textbf{ X}. The locations and shapes of artificial masks varied from each imagery. 
\begin{enumerate}

 \item Simulate a two-dimensional Gaussian random field using a covariance function cov(s, s') depending on the spatiotemporal distance between pairs of observations at locations s, s' $\in$ [0, 1].

 \item Create a binary mask \textbf{M\textsubscript{x}} by applying a threshold $\theta$ on the simulated fields.

 \item Mask corresponding pixels from \textbf{Y} to create \textbf{X} by calculating the combined mask \textbf{X = Y }$\times$\textbf{ M\textsubscript{x}}.
\end{enumerate}

The exponential covariance function $\text{cov}(s, s') = \sigma^2 \exp\left(-\frac{||s - s'||}{d}\right)$ were used, where variance $\sigma$\textsuperscript{2} = 0.95 and spatial range d=0.4 for step 1. A ratio of 10\% of pixels was selected for masking in \textbf{M\textsubscript{x}}. The generated \textbf{X} and \textbf{M\textsubscript{x}} are the inputs to our model, and \textbf{Y} is the target data used for training and testing. 

For Stage 3, six deep learning models were developed for data imputation of SD imagery on four lakes and examined their performance with the baseline model (linear interpolation). The six deep learning models include CNN, Inception Resnet, Autoencoder, CNN-LSTM, Inception Resnet-LSTM, Autoencoder -- LSTM and Resnet - LSTM. 
The dataset was split into 55\% training, 25\% validation and 20\% testing. A custom loss function, masked mean squared error (masked MSE), was implemented, calculating the MSE only for artificially masked values (Equation~\eqref{eq:loss_function}). The masked MSE was monitored in training to assess the model performance.
\begin{itemize}
  \item Let \textbf{$Y_{\text{true}}$} $\in \mathbb{R}^{H \times W \times C}$: the observed data.
  \item \textbf{$Y_{\text{pred}}$} $\in \mathbb{R}^{H \times W \times C}$: the predicted data.
  \item \textbf{$VM_{\text{pred}}$} $\in \{\mathbf{0},\mathbf{1}\}^{H \times W \times C}$: the predicted data.
  \item \textbf{$1_{\lnot \text{NaN}(Y_{i,j,c}^{\text{pred}})}$}: an indicator function that is 1 where \textbf{$Y_{\text{pred}}$} is not NaN.
\end{itemize}
Then the masked MSE loss is defined as:
\begin{equation}
\mathcal{L}_{\text{masked\_MSE}} = 
\frac{
\sum_{i,j,c} \mathrm{VM}_{i,j,c} \cdot \mathbf{1}_{\neg \text{NaN}(Y^{\text{pred}}_{i,j,c})} \cdot (Y^{\text{true}}_{i,j,c} - Y^{\text{pred}}_{i,j,c})^2
}{
\sum_{i,j,c} \mathrm{VM}_{i,j,c} \cdot \mathbf{1}_{\neg \text{NaN}(Y^{\text{pred}}_{i,j,c})}
}
\label{eq:loss_function}
\end{equation}
where:
\begin{itemize}
  \item i, j are pixel indices (height and width).
  \item c is the channel index (i.e., spectral bands).
  \item \textbf{$VM_{\text{i,j,c}}$}=1, if pixel (i,j,c) is masked (i.e., needs to be imputed), and 0 otherwise.
\end{itemize}

We used Keras (TensorFlow)\cite{abadi2016tensorflow} for implementing the deep learning models and open-source vision and image processing libraries in Python. We used a Linux computer powered by an NVIDIA A100 GPU. Each deep learning model was trained for a maximum of 50 epochs with early stopping using the Adam optimiser \cite{kingma2014adam} (learning rate of 0.001), with the custom loss function, masked MSE (Equation~\eqref{eq:loss_function}). We ran 30 independent model training experiments with different weight initialisations that ensure the robustness of the selected architecture and help quantify the variance due to different initial conditions. The deep learning models with the optimal hyperparameter configuration were trained to reconstruct \textbf{Y} using \textbf{X} and the corresponding mask \textbf{VM\textsubscript{x}}. Once trained, the models were then applied to predict the missing values in the gaps $\hat{Y}$. Meanwhile, a simple linear interpolation was implemented to generate the imputed imagery as the baseline to compare against deep learning methods. 

In Stage 4, prediction errors were compared in terms of Pearson's correlation coefficient (R), and root mean square error (RMSE) for model evaluation of the observed and imputed values in the masked pixels across eight bands (Equation~\eqref{eq:RMSE}). The smaller the RMSE values, the better the prediction accuracy: 
\begin{equation}
\text{RMSE} = \sqrt{ \frac{1}{n} \sum_{i=1}^{n} (y_i - \hat{y}_i)^2 }
\label{eq:RMSE}
\end{equation}

where, ${n}$ is the number of images, ${y}$ and $\hat{y}$ are the observed and predicted values, respectively. 

In Stage 5, we derived water quality indices from the imputed imagery within the lakes and compared them with the observed indices. The imputed imagery can be used for aquatic applications, for example, water quality index imputation. Several empirical indices \cite{brezonik2005landsat, mishra2012normalized, ho2017using} have been developed for satellite remote sensing to indicate algal bloom intensity. Algal bloom indices, such as the green/red and the normalised difference chlorophyll index (NDCI) \cite{mishra2012normalized} ($\frac{Red - Red Edge}{Red + Red Edge}$), are commonly used for algal bloom detection in lakes. For each lake, the imputed imagery from various models was extracted based on the lake polygons. The indices were applied at each pixel within the lake polygon on each day. We then compared the derived indices from the given deep learning models with the observed indices to evaluate the spatial agreement of algal levels in the masked pixels. We spatially averaged the extracted indices to generate the daily time series for each algorithm to evaluate the temporal dynamics.                  

\begin{table*}
\centering
\small
\caption{Optimal hyperparameter configurations for all six deep learning models used in this study.}
\label{tab:hyperparams}
\begin{tabular}{|l|l|l|l|l|l|}
\hline
\textbf{Model} & \textbf{Filters} & \textbf{Kernel} & \textbf{Pooling} & \textbf{LSTM units} & \textbf{Dropout} \\
\hline
CNN & 32, 64, 128 & 3$\times$3 & 2$\times$2 & -- & 0.2 \\
\hline
Autoencoder-CNN & 32, 64, 128 & 3$\times$3 & 2$\times$2 & -- & 0.2 \\
\hline
Inception-ResNet-CNN & Base model & 3$\times$3 & Global avg. & -- & 0.3 \\
\hline
CNN-LSTM & 32, 64 & 3$\times$3 & 2$\times$2 & 64 & 0.2 \\
\hline
Autoencoder-LSTM & 32, 64 & 3$\times$3 & 2$\times$2 & 64 & 0.3 \\
\hline
Inception-ResNet-LSTM & Base model & 3$\times$3 & Global avg. & 64 & 0.3 \\
\hline
\end{tabular}
\end{table*}

\section{Results}

\subsection{Model performance}

Six deep learning models (CNN, Inception ResNet, Autoencoder, CNN-LSTM, Inception-ResNet-LSTM, Autoencoder-LSTM) were evaluated for eight-band PSB.SD data imputation across four lakes in Southeastern Australia. We evaluated the temporal changes in the respective lakes over selected periods, as shown in Table~\ref{tab:lake}. 

\begin{figure*} [htbp!]
\centering
\includegraphics[width=18 cm]{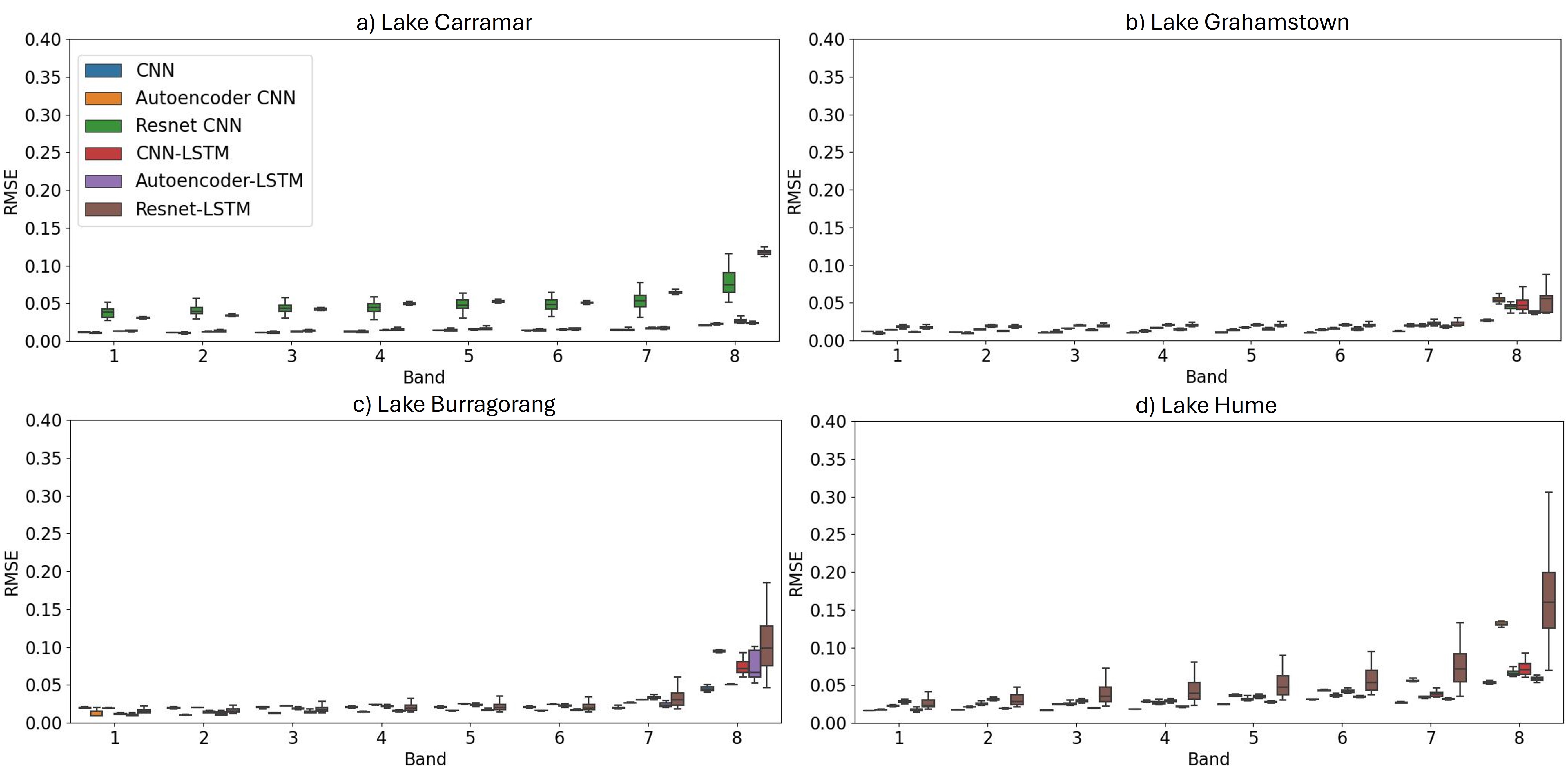}
\caption{ Model performance (RMSE) of 8 bands using CNN, Autoencoder-CNN, Inception-ResNet-CNN, CNN-LSTM, Autoencoder-LSTM and Inception-ResNet-LSTM for (a) Lake Carramar, (b) Lake Grahamstown, (c) Lake Burragorang and (d) Lake Hume. }
\label{fig:Model_performance}
\end{figure*}

We applied artificial masks generated from a Gaussian random field with a 10\% pixels mask rate to simulate cloud-masked missing pixels. The optimal configurations for each deep learning model are summarised in Table~\ref{tab:hyperparams}. The model performance was assessed across 30 independent experiments, with results reported in Table~S1. Metrics include the mean RMSE and standard deviation across all eight spectral bands for training and testing datasets, evaluated separately for each lake. 
Figure~\ref{fig:Model_performance} illustrates the model performance of the six deep learning models across four lakes for eight spectral bands. In Lake Carramar, for example, the CNN model achieved average RMSE values ranging from 0.11 to 0.21 across the eight bands (Figure~\ref{fig:Model_performance}a, Table S2). The Autoencoder model showed comparable performance, with RMSE values ranging from 0.11 to 0.23. The Autoencoder-LSTM model ranked third, with slightly higher RMSE values ranging from 0.14 to 0.25. In contrast, the Inception-ResNet-CNN and Inception-ResNet-LSTM models performed less effectively, with higher RMSE values across most bands. Notably, the Inception-ResNet-CNN model also showed greater variability in RMSE, indicating less stable performance compared to the other models. 

In Lake Grahamstown, the CNN outperformed all others, achieving RMSE values between 0.012 and 0.026 (Figure~\ref{fig:Model_performance}b, Table~S1). The Autoencoder and Autoencoder-LSTM also performed well across spectral bands 1-7, though their RMSE values were higher than those of CNN in band 8. In Lake Hume, CNN and Autoencoder-LSTM showed strong performance, with RMSE values ranging from 0.016 to 0.54 and 0.017 to 0.061, respectively (Figure~\ref{fig:Model_performance}d, Table S2). For Lake Burragorang, the Inception-ResNet-CNN and Inception-ResNet-LSTM achieved the lowest RMSE among six models; however, their performance in bands 7 and 8 (Red edge and NIR) was lower than that of CNN (Figure~\ref{fig:Model_performance}c, Table~S1), indicating limitations in handling higher-wavelength bands. Overall, CNN with optimal configurations delivered the lowest average RMSE across all eight bands in Lake Carramar, Lake Grahamstown and Lake Hume, outperforming the other five models. Autoencoder-LSTM ranked second in overall performance. Notably, RMSE tended to increase with band wavelength, with band 8 (NIR) consistently showing the highest RMSE. This trend reflects the higher variability and intensity of reflectance values in longer-wavelength bands, which pose challenges for accurate reconstruction.

\begin{figure*} [htbp!]
\centering
\includegraphics[width=18 cm]{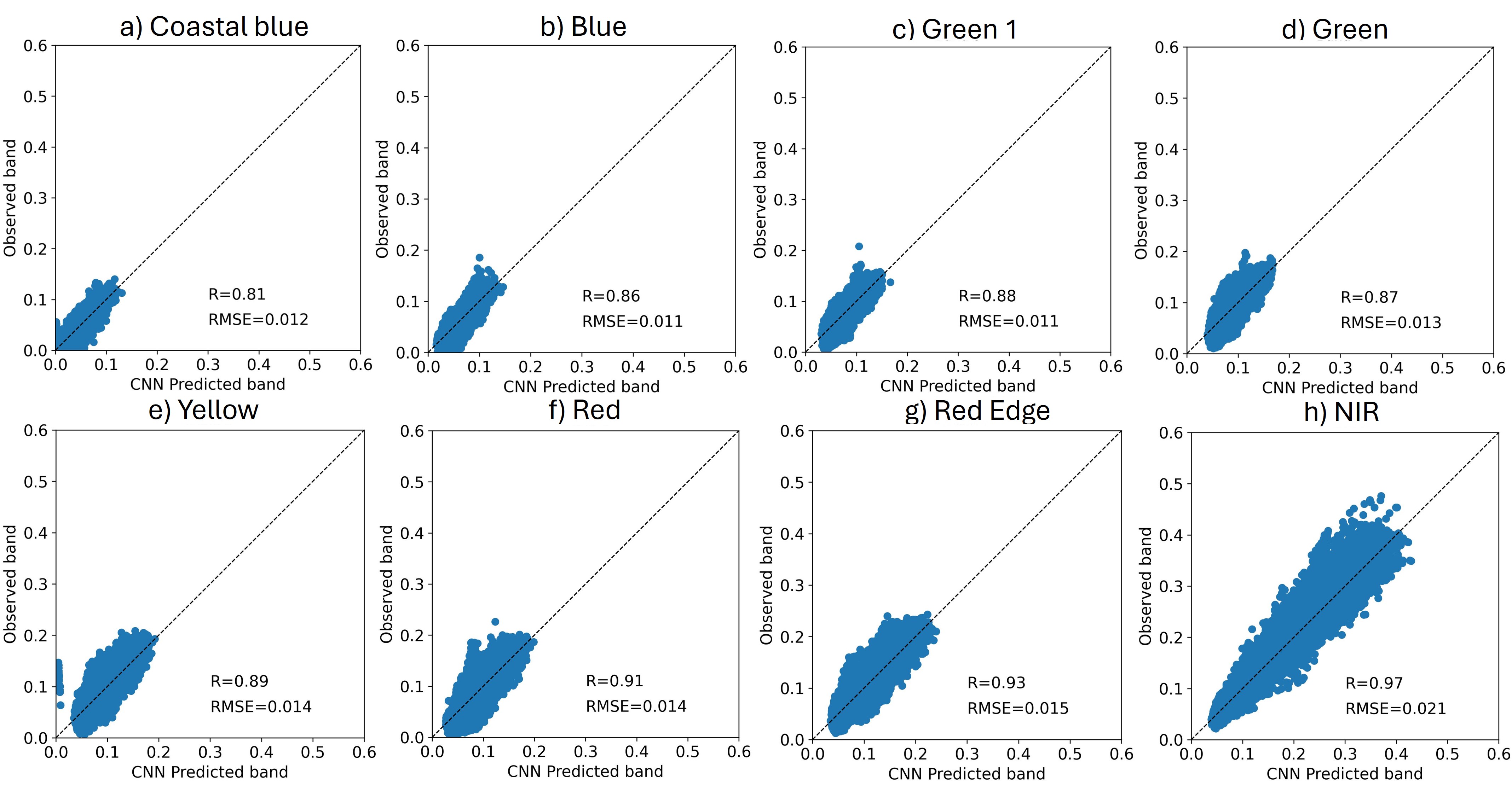}
\caption{ Comparison of CNN predicted or imputed band and observed band in the artificial masked pixels for (a) coastal blue (band 1), (b) blue (band 2), (c) green 1 (band 3), (d) green (band 4), (e) yellow (band 5), (f) red (band 6), (g) red edge (band 7) and (g) NIR (band 8) of Lake Carramar. }
\label{fig:scatterplot}
\end{figure*}

Figure~\ref{fig:scatterplot} presents scatter plots comparing predicted and observed values for artificially masked pixels across eight bands. The CNN model demonstrated strong performance, with high Pearson correlation (R) and low RMSE values. For example, in Lake Carramar, R values ranged from 0.81 to 0.93, spanning from coastal blue (band 1) to NIR (band 8). RMSE values remained below 0.015 for coastal blue through red edge bands, with a slightly higher RMSE of 0.021 for NIR. These results confirmed the CNN model's effectiveness in accurately reconstructing missing data. Additional scatter plots for Lake Grahamstown, Lake Burragorang and Lake Hume are provided in the Supplementary material Figure S2-4, further illustrating the consistency of CNN performance across different lakes.

\begin{figure*} [htbp!]
\centering
\includegraphics[width=15 cm]{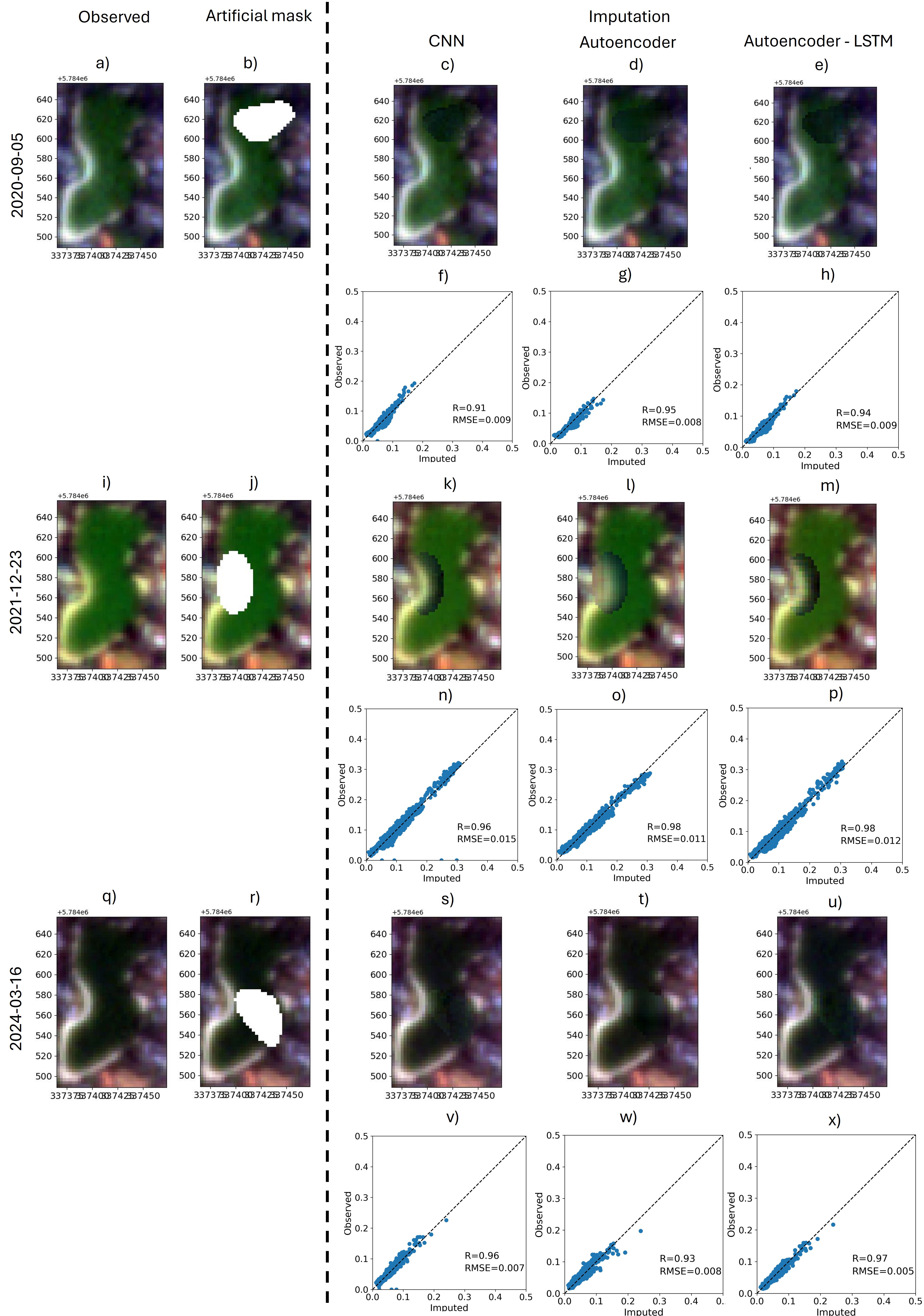}
\caption{ RGB for (a) observed, (b) additional cloud mask (in white), and (c-e) top three best models (CNN, Autoencoder-CNN, Autoencoder-CNN-LSTM) imputed imagery of Lake Carramar imagery on 5th September 2020. Comparison of the observed and imputed values of 8 bands in the masked pixels for (f) CNN, (g) Autoencoder-CNN, and (h) Autoencoder-LSTM, with R and RMSE shown in each plot. (i-p) and (q-x) is the same as (a-h) but for 23rd December 2021 and 16th March 2024. Note that the location, size and shape of the artificial clouds varied for individual days. }
\label{fig:Carramar_image}
\end{figure*}

In the earlier evaluation (Table~S1), CNN, Autoencoder-CNN and Autoencoder-CNN-LSTM emerged as the top three performing models. To visualize the reconstructed data, we presented observed and imputed imagery for Lake Carramar (Figure~\ref{fig:Carramar_image}) over three distinct dates, each representing different water quality conditions: low algal blooms (2020-09-05, Figure~\ref{fig:Carramar_image} a-e), high algal blooms (2021-12-23, Figure~\ref{fig:Carramar_image} f-j), and clear water conditions (2024-03-06, Figure~\ref{fig:Carramar_image} k-o). Artificial masks were applied to simulate cloud cover, mainly affecting water pixels and occasionally extending to the lake's edges. The shape and location of these masks varied across dates to reflect realistic cloud patterns. All three models produced imputed imagery that closely matched the observed data within the masked regions (Figure~\ref{fig:Carramar_image}), demonstrating strong spatial consistency. Specifically, the CNN model achieved R between 0.91 and 0.96, and RMSE between 0.007 and 0.015 within the masked pixels (Figure~\ref{fig:Carramar_image} c, k, s). The Autoencoder-CNN (Figure~\ref{fig:Carramar_image} d, l, t) and Autoencoder-LSTM (Figure~\ref{fig:Carramar_image} e, m, u) also showed high performance, with similarly strong agreement between imputed and observed values.

\begin{figure*} [htbp!]
\centering
\includegraphics[width=20 cm]{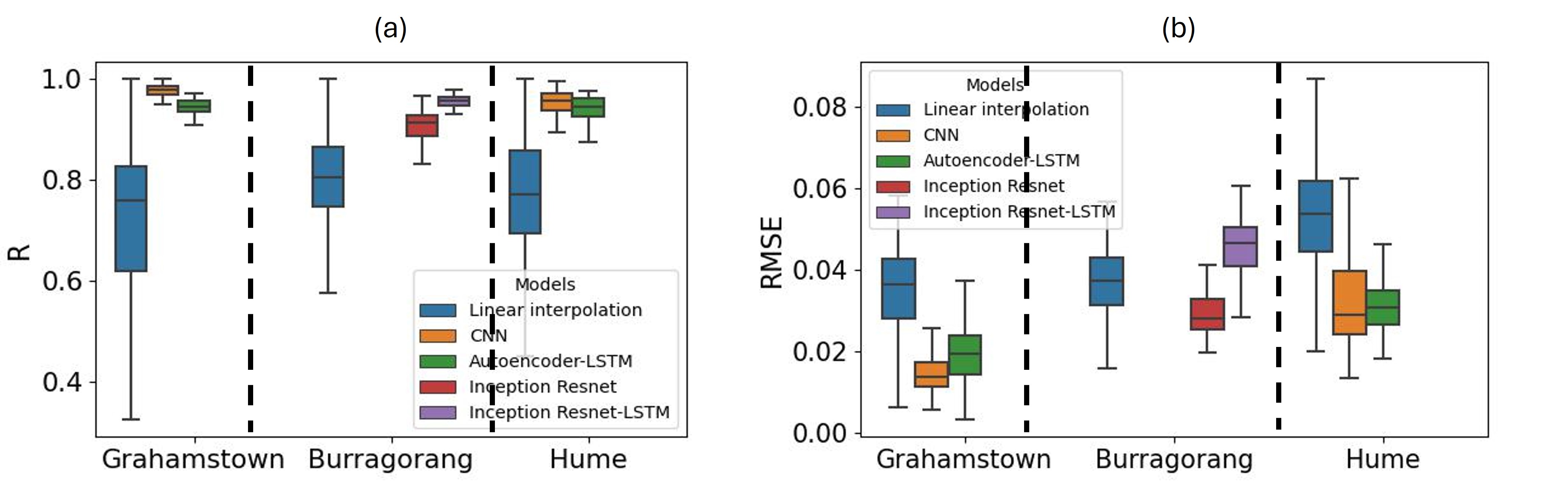}
\caption{ The comparison of (a) R and (b) RMSE of the observed against the imputed values of 8 bands for linear interpolation (in blue) and the top two deep learning models for Lake Grahamstown, Burragorang and Hume across 4 years. Note that the top two deep learning models for Lake Grahamstown and Hume are CNN (in orange) and Autoencoder-LSTM (in green), while the top two models for Burragorang are Inception-ResNet (in red) and Inception-ResNet-LSTM (in purple). }
\label{fig:three_lakes_comparison}
\end{figure*}

Figure~\ref{fig:three_lakes_comparison} illustrates the improvement in imputation across eight spectral bands within masked regions by comparing the baseline linear interpolation model with the top-performing deep learning models for Lake Grahamstown, Lake Burragorang and Lake Hume, over the period from August 2020 to December 2024. For Lake Grahamstown and Hume, CNN and Autoencoder-LSTM were identified as the best performing models, while for Lake Burragorang, Inception-ResNet and Inception Resnet-LSTM showed better results. Across all three lakes, the deep learning models consistently achieved higher median R (R > 0.8) compared to linear interpolation, which also showed greater variability in R values (Figure~\ref{fig:three_lakes_comparison} a). For Lake Grahamstown and Hume, CNN and Autoencoder-LSTM (median RMSE below 0.02) outperformed linear interpolation (Figure~\ref{fig:three_lakes_comparison} b). In Lake Burragorang, the Inception-ResNet model (median RMSE: 0.027) outperformed the linear interpolation (median RMSE: 0.038) and the Inception-ResNet-LSTM (median RMSE: 0.048). These results highlight the robustness and adaptability of deep learning models for reconstructing missing data across diverse lake types.

\begin{figure*} [htbp!]
\centering
\includegraphics[width=17 cm]{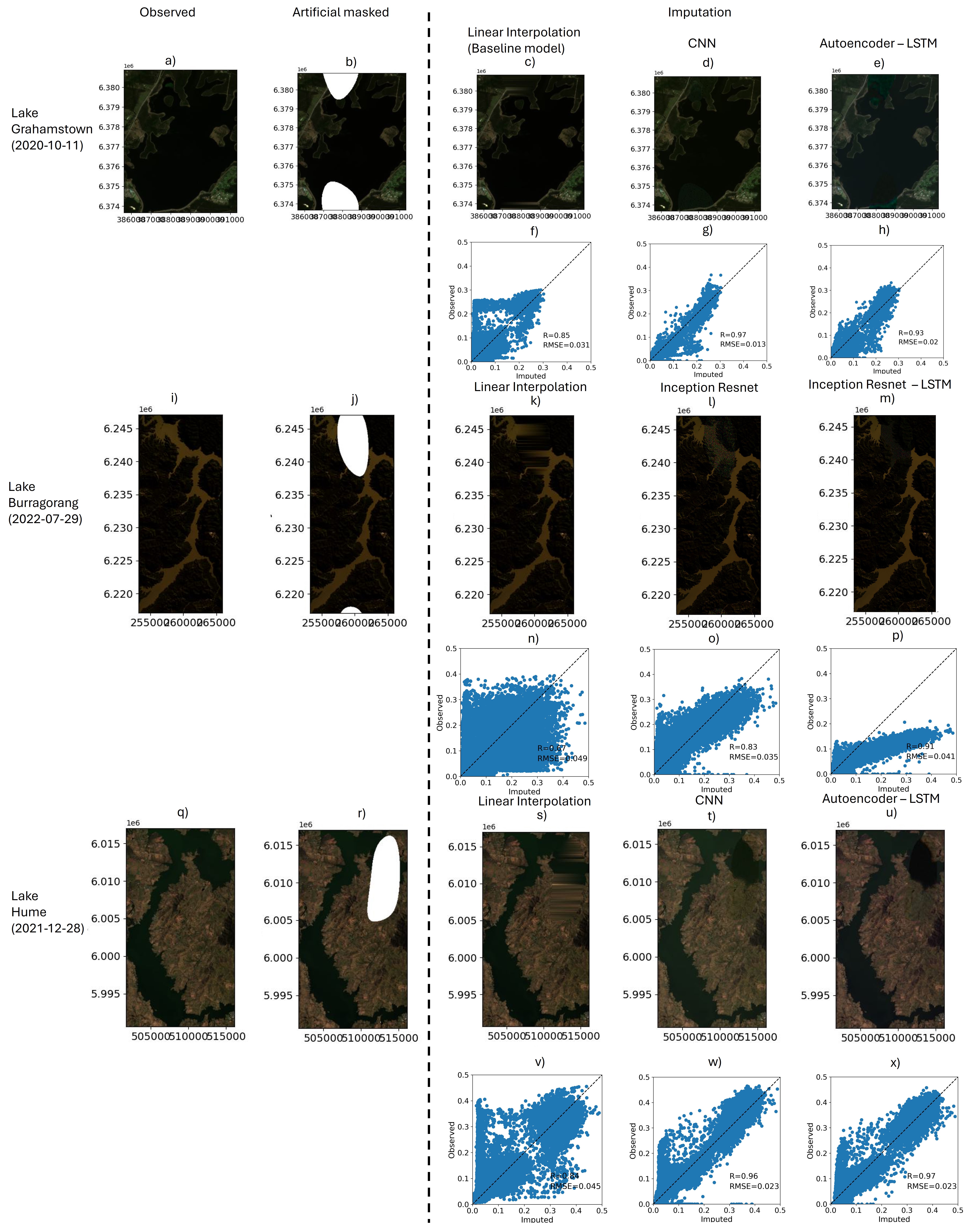}
\caption{ RGB image data for (a) observed, (b) additional cloud masked (in white), (c) linear interpolation (as baseline model) imputed and the top two best models, (d) CNN and (e) Autoencoder-LSTM imputed imagery for Lake Grahamstown (11th October 2020). We provide a comparison of the observed and imputed values of 8 bands in the masked pixels for (f) linear interpolation, (g) CNN and (h) Autoencoder-LSTM, with R and RMSE shown in each plot. (i-p) and (q-x) is the same as (a-e), but for Lake Burragorang (29th July 2022) and Lake Hume (28th December 2021). Inception ResNet and Inception-ResNet--LSTM are the top two best models for Lake Burragorang, while CNN and Autoencoder--LSTM make the top two best models for Lake Hume. Note that the location, size and shape of the generated artificial clouds varied for individual days. }
\label{fig:three_lakes_image}
\end{figure*}

To visualise the imputed pixels, one representative date was selected for each lake. The deep learning models consistently outperformed the baseline linear interpolation method, showing superior spatial agreement and improved performance metrics across all eight spectral bands within the masked areas. For Lake Grahamstown, linear interpolation yielded a correlation of 0.85 with an RMSE of 0.031 (Figure~\ref{fig:three_lakes_image}c, f). In contrast, CNN and Autoencoder-LSTM achieved R above 0.93 and RMSE below 0.02 (Figure~\ref{fig:three_lakes_image}d-e, g-h), indicating more accurate imputation. Similarly, in Lake Hume, the deep learning models achieved correlations above 0.96 and an RMSE of 0.023 (Figure~\ref{fig:three_lakes_image}t-u, w-x), compared to linear interpolation (R\~0.84, RMSE\~0.045) (Figure~\ref{fig:three_lakes_image}s, v). In Lake Burragorang, the top-performing models, Inception-ResNet and Inception-ResNet--LSTM, also provided better imputation results than linear interpolation (Figure~\ref{fig:three_lakes_image}k-p). However, the Inception-ResNet--LSTM tended to overestimate the band values (Figure~\ref{fig:three_lakes_image}p), indicating limitations in its temporal modelling for this lake.

\subsection{Application of imputed imagery - water quality indices as a case study}

\begin{figure*} [htbp!]
\centering
\includegraphics[width=17 cm]{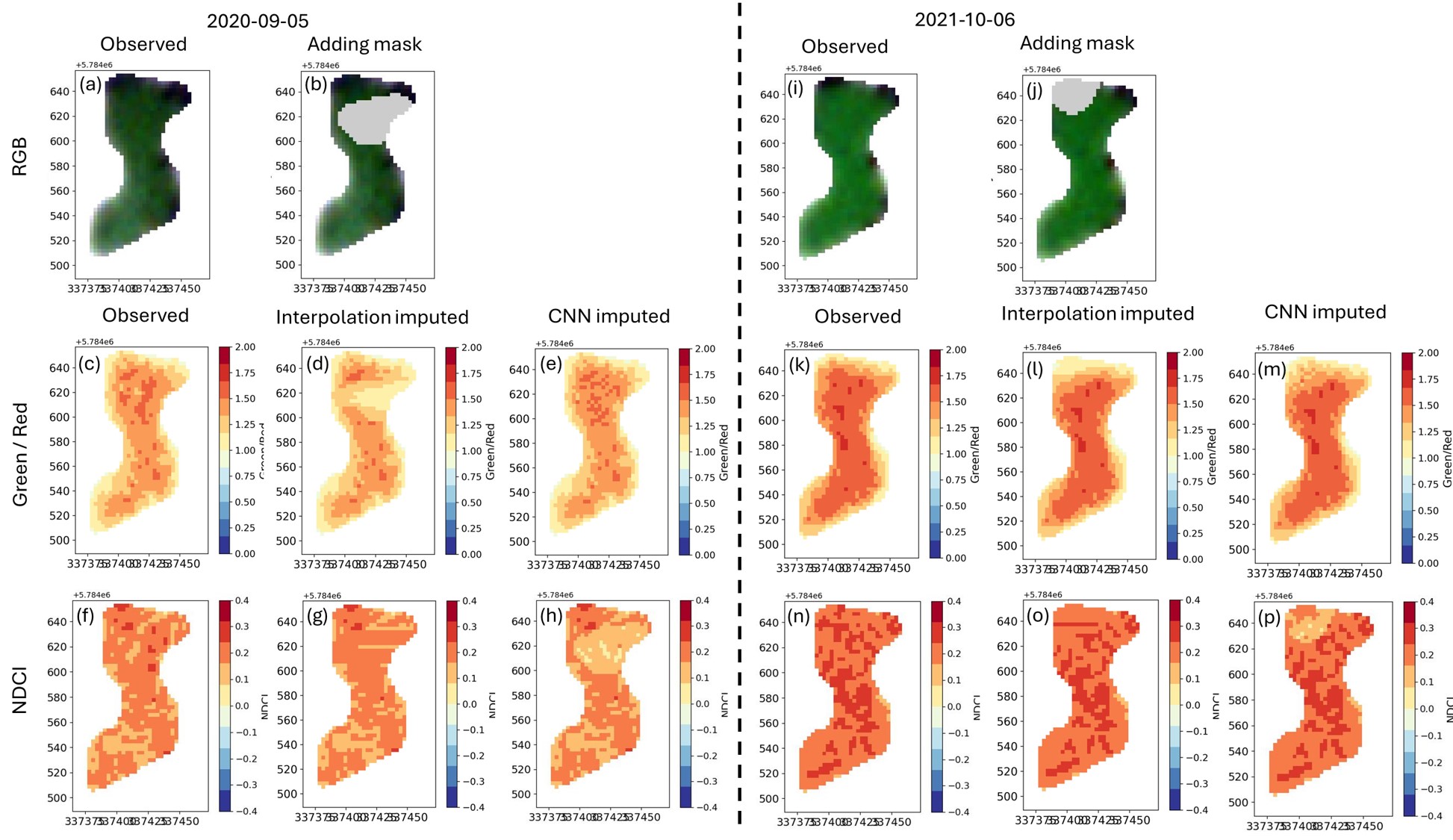}
\caption{ We present Lake Carramar in 5th September 2020 as an example using linear interpolation and CNN imputation derived NDCI with Panel (a) RGB, Panel (b) additional cloud mask (in grey), Panel (c) observed, Panel (d) linear interpolation imputed, Panel (e) CNN imputed derived Green/Red, and Panel (f-h) as the observed. Panel (i-p) presents the same region as Panel (a-h) but for date 6 October 2021. Note that the location, size and shape of the artificial clouds varied for individual days. }
\label{fig:Cammarar_indices}
\end{figure*}

\begin{figure*} [htbp!]
\centering
\includegraphics[width=17 cm]{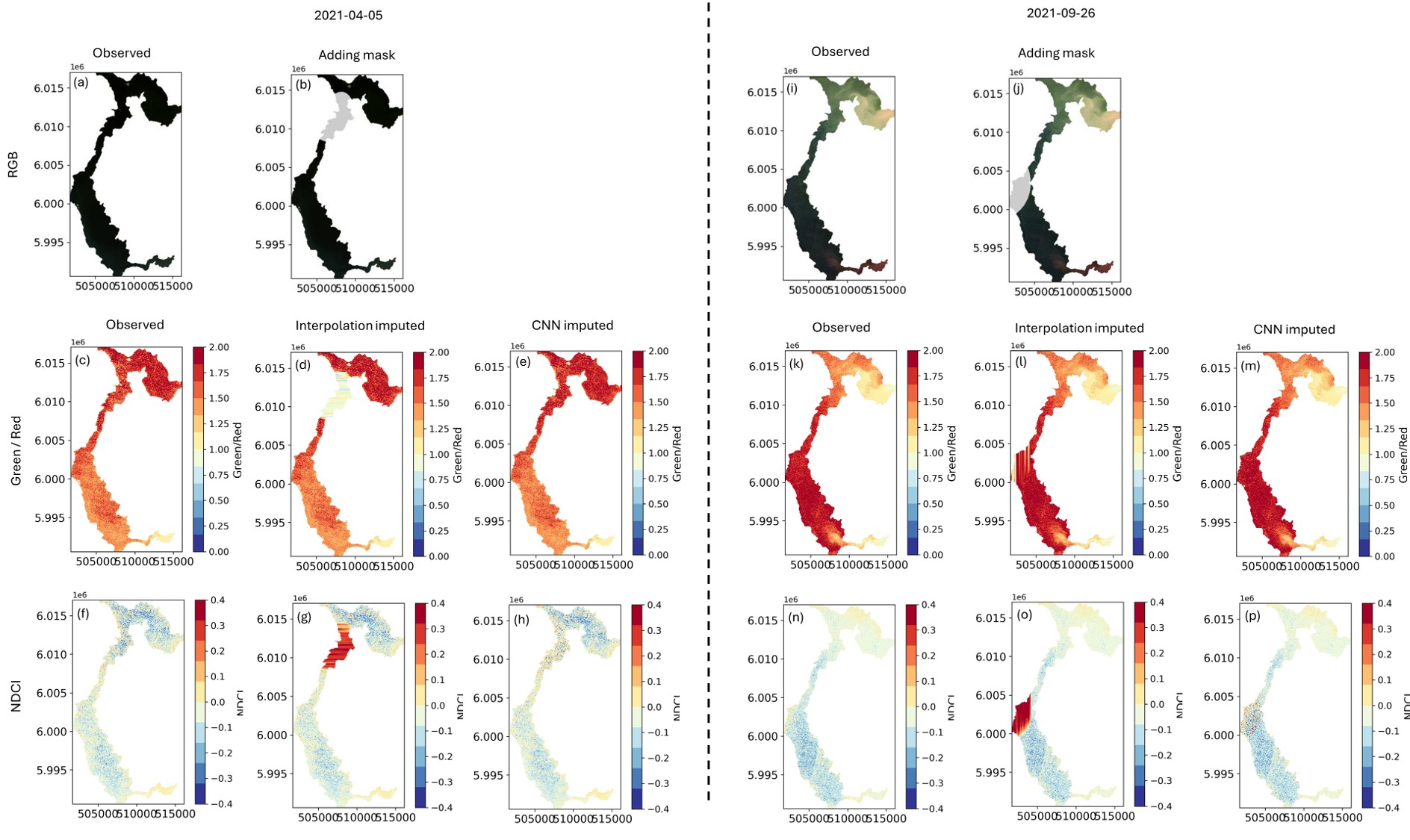}
\caption{ We present Lake Hume in 5th April 2021 as an example using linear interpolation and CNN imputation derived NDCI with Panel (a) RGB, Panel (b) additional cloud mask (in grey), Panel (c) observed, Panel (d) linear interpolation imputed, Panel (e) CNN imputed derived Green/Red, and Panel (f-h) as the observed. Panel (i-p) presents the same region as Panel (a-h), but for the date 26 September 2021. Note that the location, size and shape of the artificial clouds varied for individual days. }
\label{fig:Hume_indices}
\end{figure*}

The imputed imagery enables further exploration of both terrestrial (e.g., vegetation and land cover) and aquatic monitoring (e.g., water extent and water quality). In this study, we focus on water quality assessment using indices derived from the imputed PSB.SD imagery. The indices, such as the Green/Red ratio and the Normalised Difference Chlorophyll-a Index (NDCI) \cite{mishra2012normalized}, are commonly calculated from specific spectral bands and serve as indicators for algal concentrations in lakes. Lake Carramar and Lake Hume have experienced frequent algal blooms in recent years, making them suitable for evaluating the effectiveness of imputed imagery in algal bloom monitoring. To visualise the spatial distribution of water quality indices within the masked lake pixels, we compared observed data, linear interpolation results and CNN imputation for both the Green/Red and NDCI. The results for Lake Carramar and Hume are presented in Figures~\ref{fig:Cammarar_indices} and~\ref{fig:Hume_indices}, respectively, using two representative dates for each lake. 

The Green/Red and NDCI imputation from linear interpolation for Lake Carramar exhibited a noticeable smoothing effect in the masked regions (Figure~\ref{fig:Cammarar_indices}d, l, g, o). This smoothing is consistent with patterns observed in the individual spectral bands (Figure~S1), reflecting the limitations of linear interpolation in capturing spatial detail. In contrast, CNN imputation demonstrates strong spatial agreement with the observed data in the masked pixels for Green/Red. However, the CNN-imputed NDCI values tended to be lower than the observed NDCI values in the masked areas, suggesting an underestimation. This discrepancy may arise from how NDCI was calculated - using a combination of selected spectral bands - where CNN achieved lower RMSE in the Green and Red bands, but higher RMSE in the Red Edge band, which contributes to NDCI. A similar pattern is observed in Lake Hume, linear interpolation results in low spatial agreement in both indices (Figure~\ref{fig:Hume_indices}), while CNN-imputed Green/Red and NDCI values show high spatial consistency with the observed data. However, on both selected dates, CNN slightly overestimated the values of these indices, indicating a minor bias in the imputation. 

\begin{figure*} [htbp!]
\centering
\includegraphics[width=13 cm]{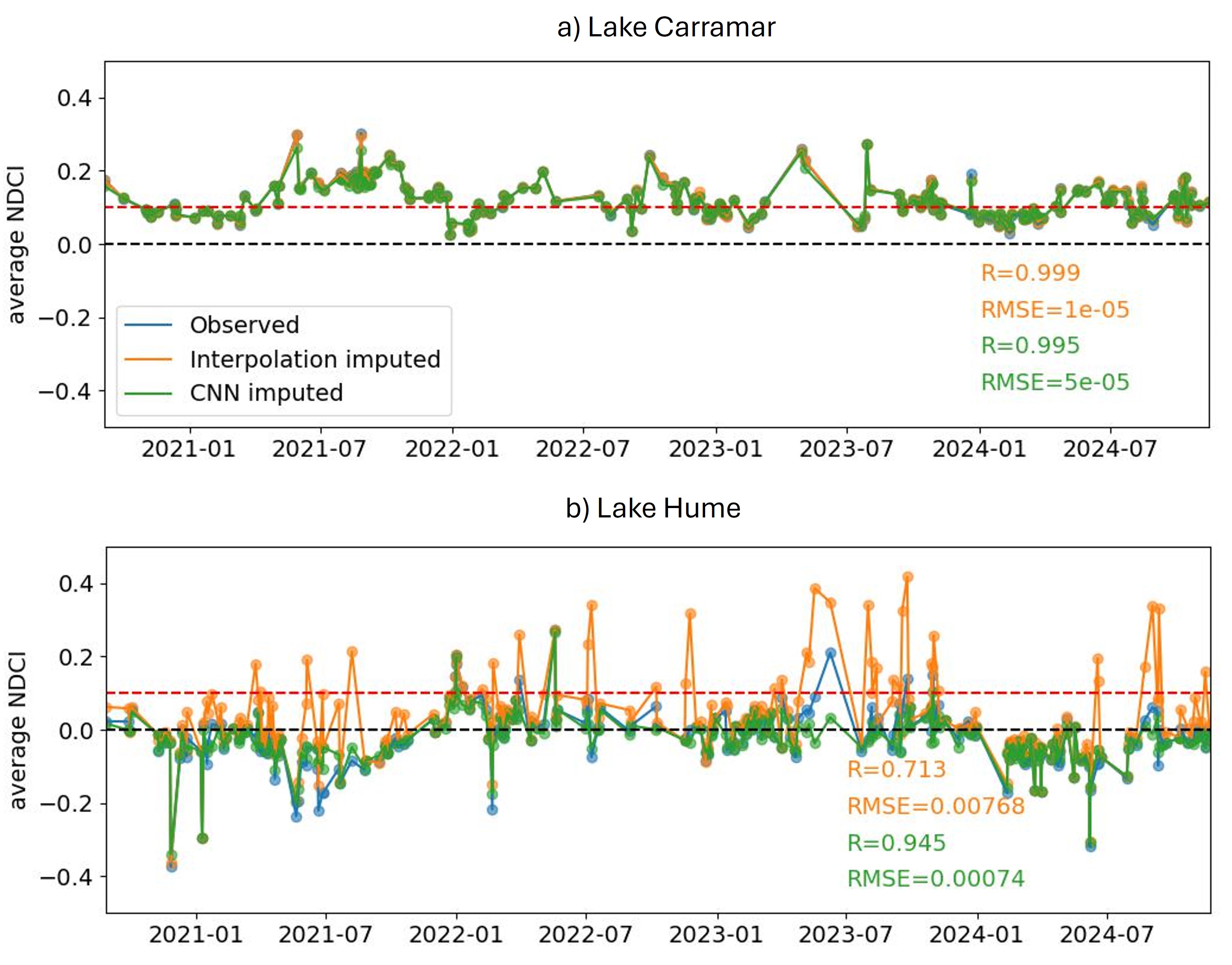}
\caption{Timeseries of average NDCI extracted by the observed (in blue), linear interpolation imputed (in orange) and CNN model imputed (in green) for (a) Lake Carramar and (b) Lake Hume from August 2020 to December 2024. Pearson correlation R and RMSE of the observed and the imputed average NDCI are shown in (a) and (b). The black dashed line indicates the threshold of low algal biomass (NDCI<0), and the red dashed line indicates the threshold of algal bloom risk (NDCI>0.1).
}
\label{fig:timeseries_indices}
\end{figure*}

We generated time series data of average NDCI values across lake pixels using the imputed PSB.SD imagery for Lake Carramar and Lake Hume (Figure~\ref{fig:timeseries_indices}). In Lake Carramar, the observed average NDCI consistently remained above zero, with noticeable fluctuations over time, indicating the presence of chlorophyll-a (Chla) and potential algal bloom events. Both the linear interpolation and CNN models closely matched the observed average NDCI values, achieving high R values above 0.995 (Figure~\ref{fig:timeseries_indices}a). In contrast, Lake Hume exhibited more dynamic temporal patterns. From August 2020 to September 2021, the observed NDCI values were around or below 0, suggesting low Chla levels. From October 2021 to December 2023, NDCI values increased, indicating elevated Chla concentrations, before declining again below zero through December 2024 (Figure~\ref{fig:timeseries_indices}b). The CNN model effectively captured these temporal trends, achieving an R of 0.94 and an RMSE of 0.001, outperforming linear interpolation (R=0.71, RMSE=0.008), which showing a tendency to overestimate NDCI values on multiple days. 
NDCI values serve as indicators of algal concentration in lakes. Values below 0 indicate low algal levels, values between 0 and 0.1 indicate moderate to high algal levels, and values greater than 0.1 signal an algal bloom risk. Based on these thresholds, we classified each day into three categories: low algal levels, moderate to high algal levels, and algal bloom risk. For Lake Carramar, none of the days fell into the low algal level category. Approximately 43.5\% of the days were classified as moderate to high algal levels, while 56.5\% were identified as being at risk of algal blooms. These classifications, derived from both linear interpolation and CNN-imputed imagery, were consistent with the observations, confirming the reliability of the imputation methods for water quality assessment. 
For Lake Hume, 62\% of the observed days were classified as low algal levels (NDCI<0), 34.6\% showed moderate to high algal levels (0<NDCI<0.1), and 33.8\% were identified as being algal bloom risk (NDCI>0.1). When using linear interpolation, 44.0\% of days were estimated as low algal levels (NDCI<0), whereas CNN imputation classified 70.1\% of days in this category, closer to the observed. For moderate to high algal levels (0<NDCI<0.1), linear interpolation estimated 52.6\% of days, while CNN imputation identified 25.9\%. For algal bloom risk, both models showed good agreement with observations. Overall, CNN imputation demonstrated higher classification accuracy for Lake Hume compared to linear interpolation. 

\section{Discussion}

Cloud cover and satellite technical limitations are common challenges in the application of optical satellites for water quality monitoring. In this study, we used imagery from the PlanetScope SuperDove satellite as a case study to evaluate the effectiveness of data imputation across all spectral bands within artificial masked lake regions. These artificial masks were designed to simulate realistic cloud cover, which frequently obscures satellite observations. We evaluated the imputation performance of six deep learning models against a baseline linear interpolation method. All deep learning models achieved relatively low RMSE values across eight spectral bands for four lakes, with slightly higher RMSE values in the NIR band. For instance, in Lake Carramar, the CNN model achieved R ranging from 0.83 to 0.97, and RMSE values between 0.011 and 0.021, respectively. CNN consistently delivered the best performance (i.e., lowest RMSE) for Lake Carramar, Lake Grahamstown, and Lake Hume, while the Inception Resnet model performed best for Lake Burragorang. This discrepancy may be attributed to the differences in the proportion of water pixels in each image, influenced by the shape and irregularity of the lakes' boundaries. 

Across all eight spectral bands and masked regions, deep learning models consistently outperformed the baseline model. The top-performing deep learning models improved R by 0.13 -- 0.17, and reduced RMSE by 0.006 -- 0.019 compared to the baseline. Overall, CNN-based models outperformed the CNN-LSTM-based model, showing higher R and lower RMSE. Despite the nominal sub-daily to daily revisit frequency provided by the Planetscope SuperDove, the actual temporal resolution in the study lakes was often beyond 2-3 days, particularly during the first two years of the study period. Extended periods of cloud cover further contributed to consecutive days of missing imagery, resulting in irregular time series rather than consistent daily observations. Since the LSTM models were trained using sequences of five timesteps, these gaps likely disrupted temporal continuity and contributed to the reduced performance of the CNN-LSTM models compared to purely spatial CNN architectures. 

The imputed values across eight spectral bands enable the derivation of various indices for aquatic applications. In this study, we focused on algal bloom indices as a case example, given that all four lakes investigated have historical records of bloom events. However, during the study period (August 2020 -- December 2024), not all lakes experienced active algal blooms. We derived two commonly used algal bloom indices (i.e., Green/Red and NDCI) from both the imputed and original imagery to assess temporal changes. Beyond algal bloom monitoring, other empirical water quality algorithms, such as those for turbidity \cite{brezonik2005landsat,dekker2002analytical,wang2006applications}, color dissolved organic matter \cite{brezonik2005landsat, schroeder2008remote} can also be applied to the imputed data. In addition to water quality indices, other indices, such as NDVI, NDWI \cite{gao1996ndwi, mcfeeters1996use} can also be derived from the imputed data. If the primary focus is on a specific index, several approaches can be applied. One approach involves extracting the target index as one channel and applying deep learning models directly to impute the missing values for that index. Alternatively, principal component analysis can be used to identify key spectral components, which can then be processed using CNN or other deep learning techniques to enhance imputation accuracy. Moreover, the methodology presented in this study can be generalised to other satellites, including hyperspectral sensors. This flexibility allows for broader applications in environmental monitoring, where high-resolution and multi-band data are essential for accurate analysis and prediction. 

Additionally, the imputed imagery can support more advanced analytical and machine learning approaches \cite{chen2022remote,sagan2020monitoring} for robust monitoring and prediction of algal blooms. As expected, the algal bloom indices derived from deep learning-imputed imagery generally outperformed those from the baseline model. For instance, NDCI derived from CNN-imputed imagery for Lake Hume showed closer alignment with the observed data, with an average R increasing by 0.23 and RMSE decreasing by 0.007 compared to the baseline. Algal blooms are highly variable in space and time \cite{liu2022remote}, and often comprise different dominant species such as phytoplankton, diatoms or cyanobacteria. These blooms can move horizontally and vertically within the water column \cite{liu2021effectiveness}, and their duration can vary from a few days to weeks. However, this study focused on imagery with scattered cloud cover, excluding full cloud-covered scenes. Therefore, the training dataset may lack sufficient representation of peak algal bloom conditions, potentially limiting the model's predictive performance for high algal concentrations. To address this limitation, data augmentation techniques, such as upsampling, could be useful to simulate high algal concentration scenarios. This would help ensure the training dataset better reflects peak bloom conditions and improve model robustness in imputing intense bloom events.

Moreover, the NDCI threshold used to classify algal blooms can vary between lakes due to differences in their ecological and optical characteristics. In this study, we applied a common threshold range - NDCI <0, 0-0.1, and >0.1 - to classify water conditions into low, moderate and algal bloom levels. Classification performance using CNN-imputed imagery showed improvements over linear interpolation. However, given the unique characteristics of each lake, the optimal NDCI threshold may differ across lakes. For instance, \cite{caballero2021monitoring} showed a NDCI threshold of 0.2 for algal bloom identification for Lake Laguna, while \cite{zhi2024heterogeneity} and \cite{aubriot2020assessing} adopted a NDCI threshold of 0.06 for the Three Gorges Reservoir and Palmar reservoir. To accurately identify regions affected by algal blooms, validation using in-situ data or outputs from a water quality model is required. On the other hand, satellite imagery, including imputed data, could complement field observations, especially in areas with limited monitoring, by enhancing both calibration and validation efforts. The use of imputed imagery to fill gaps caused by cloud cover strengthens the continuity of water quality monitoring and prediction models, with caution directed toward the associated uncertainties. This enhancement supports improved identification of algal blooms and other water quality issues, contributing to enhanced water quality monitoring and catchment management. 

Cloud contamination and low pixel quality remain challenges for optical satellite applications, particularly for water quality monitoring. In data-driven approaches, the quality of the training dataset is crucial. As such, the UDM masks --- which identify the pixel conditions (e.g., clear or cloud-affected) play an important role in data preprocessing. However, in our study, inaccuracies in UDM mask classifications were observed for PlanetScope SuperDove imagery. Specifically, some pixels affected by clouds, haze or cloud shadows were incorrectly labelled as clear, allowing contaminated data into the training dataset. This misclassification potentially degraded model performance and reduced the accuracy of the imputed values. Therefore, enhancing the reliability of cloud and cloud-shadow detection during preprocessing is essential for robust optical satellite analysis. While many studies have investigated cloud detection and gap-filling techniques for widely used satellites, such as MODIS, Landsat, Sentinel 2 -- benefiting from their long-term archives -- studies focused on high-resolution satellites such as PlanetScope SuperDove remain limited. Despite its advantages in spatial and temporal resolution and its spectral similarity to Sentinel 2, Planetscope SuperDove is still underexplored in this context. In this study, we assessed multiple deep learning models to impute missing data across eight spectral bands of various lake types and conditions. The imputed imagery was then used to derive water quality indices. Importantly, since our primary objective was to compare model performance, the presence of cloud contamination is unlikely to bias the relative evaluation across models. Nevertheless, to improve the overall accuracy and reliability of imputed data, further advancements in cloud cover identification for Planetscope SuperDove and similar optical satellites are necessary. Enhancing these preprocessing steps will support more effective water quality monitoring and prediction, ultimately benefiting lake and catchment management. 

In this study, we artificially applied cloud masks to 10\% of image pixels to simulate partially obscured satellite data. In real-world scenarios, however, cloud cover in satellite imagery can vary considerably. Previous studies have shown that as the percentage of cloud cover increases, the accuracy of imputed satellite data tends to decrease \cite{chen2022novel,daniels2022filling, liang2023reconstructing}. While it is possible to assess the impact of different cloud cover percentages, our findings suggest that these variations have minimal influence on the relative performance comparison between deep learning models. A key limitation of our study is that the models were only tested on imagery with partial cloud cover. Their ability to reconstruct scenes that are fully obscured by clouds was not evaluated. Assessing this limitation would require further research into model performance under more extreme cloud conditions, which is critical for advancing robustness of satellite-based monitoring in highly cloud-prone regions. 

Future study is recommended to focus on the following research gaps: 
a) Enhancing cloud and cloud-shadow detection: Enhancing the accuracy and reliability of cloud and shadow classification is essential for ensuring high-quality training data in optical satellite applications. 
b) Evaluating model performance under high cloud coverage: For Planetscope SuperDove imagery, further investigation is needed to assess the effectiveness of imputation models under conditions of high cloud cover, to improve dataset completeness in challenging scenarios. 
c) Establishing lake-specific thresholds for water quality indices: Developing tailored thresholds for indices such as NDCI through validation with in-situ measurements or outputs for a water quality model with imputed imagery;
d) Incorporating data resampling and augmentation techniques: Applying methods such as upsampling or synthetic data augmentation can improve the representation of rare or extreme events (e.g., algal blooms), thereby improving the predictive accuracy of water quality indicators.

\section{Conclusion}

Data imputation in optical remote sensing is essential for enhancing the robustness of water quality monitoring, such as algal bloom detection. While previous research has largely focused on widely used satellites such as MODIS, Landsat, Sentinel 2, this study extends data imputation techniques to the Planetscope SuperDove - a high-resolution optical satellite increasingly used in monitoring due to its superior spatial and temporal coverage. We evaluated the performance of a baseline model (i.e., linear interpolation) and deep learning models for imputing missing spectral band values across four lakes. The top-performing deep learning model for each lake consistently showed superior performance in reconstructing spectral band values in the artificial masked regions, compared to the baseline model. To assess the practical utility of the imputed imagery, we applied it to algal bloom detection using empirical algorithms such as Green/Red and NDCI. These indices derived from deep learning-imputed data closely matched those from observed imagery. Our findings demonstrated that the deep learning-based imputation substantially improves the accuracy of blooming classification compared to the baseline method, underscoring its potential to enhance satellite-based environmental monitoring.

This study evaluated the effectiveness of deep learning techniques for imputing missing values of all spectral bands in optical satellite imagery and demonstrated the practical utility of the resulting imputed images. By filling data gaps, the high-resolution Planetscope SuperDove imagery can provide additional insights for comprehensive spatiotemporal analysis of water quality. This enhancement could further enhance accurate prediction and early warning of water quality events (e.g., harmful algal blooms), while also improving risk assessment and identifying vulnerable regions. The methodology presented in this study can be generalised to other optical satellites and can be extended to a wide range of aquatic and terrestrial monitoring applications, offering valuable support for catchment and water resource management. 

\section{Data and Code availability}

Data and Python code for the models are given via GitHub repo: \url{ }.

 \bibliographystyle{elsarticle-num} 
 \bibliography{cas-refs}

@article{ho2019widespread,
  title={Widespread global increase in intense lake phytoplankton blooms since the 1980s},
  author={Ho, Jeff C and Michalak, Anna M and Pahlevan, Nima},
  journal={Nature},
  volume={574},
  number={7780},
  pages={667--670},
  year={2019},
  publisher={Nature Publishing Group UK London}
}

@incollection{matthews2017bio,
  title={Bio-optical modeling of phytoplankton chlorophyll-a},
  author={Matthews, Mark W},
  booktitle={Bio-optical modeling and remote sensing of inland waters},
  pages={157--188},
  year={2017},
  publisher={Elsevier}
}

@article{houborg2018cubesat,
  title={A cubesat enabled spatio-temporal enhancement method (cestem) utilizing planet, landsat and modis data},
  author={Houborg, Rasmus and McCabe, Matthew F},
  journal={Remote Sensing of Environment},
  volume={209},
  pages={211--226},
  year={2018},
  publisher={Elsevier}
}

@article{le2013evaluation,
  title={Evaluation of chlorophyll-a remote sensing algorithms for an optically complex estuary},
  author={Le, Chengfeng and Hu, Chuanmin and Cannizzaro, Jennifer and English, David and Muller-Karger, Frank and Lee, Zhongping},
  journal={Remote Sensing of Environment},
  volume={129},
  pages={75--89},
  year={2013},
  publisher={Elsevier}
}

@article{verpoorter2014global,
  title={A global inventory of lakes based on high-resolution satellite imagery},
  author={Verpoorter, Charles and Kutser, Tiit and Seekell, David A and Tranvik, Lars J},
  journal={Geophysical Research Letters},
  volume={41},
  number={18},
  pages={6396--6402},
  year={2014},
  publisher={Wiley Online Library}
}

@article{chawla2020review,
  title={A review of remote sensing applications for water security: Quantity, quality, and extremes},
  author={Chawla, Ila and Karthikeyan, L and Mishra, Ashok K},
  journal={Journal of Hydrology},
  volume={585},
  pages={124826},
  year={2020},
  publisher={Elsevier}
}

@article{gholizadeh2016comprehensive,
  title={A comprehensive review on water quality parameters estimation using remote sensing techniques},
  author={Gholizadeh, Mohammad Haji and Melesse, Assefa M and Reddi, Lakshmi},
  journal={Sensors},
  volume={16},
  number={8},
  pages={1298},
  year={2016},
  publisher={MDPI}
}

@article{liu2022remote,
  title={Remote sensing to detect harmful algal blooms in inland waterbodies},
  author={Liu, S and Glamore, W and Tamburic, B and Morrow, A and Johnson, F},
  journal={Science of the Total Environment},
  volume={851},
  pages={158096},
  year={2022},
  publisher={Elsevier}
}

@article{marta2018planet,
  title={Planet imagery product specifications},
  author={Marta, Santa},
  journal={Planet Labs: San Francisco, CA, USA},
  volume={91},
  pages={170},
  year={2018}
}

@article{bareuther2020spatio,
  title={Spatio-temporal dynamics of algae and macrophyte cover in urban lakes: a remote sensing analysis of Bellandur and Varthur wetlands in Bengaluru, India},
  author={Bareuther, Mischa and Klinge, Michael and Buerkert, Andreas},
  journal={Remote Sensing},
  volume={12},
  number={22},
  pages={3843},
  year={2020},
  publisher={MDPI}
}

@article{beck2017comparison,
  title={Comparison of satellite reflectance algorithms for estimating phycocyanin values and cyanobacterial total biovolume in a temperate reservoir using coincident hyperspectral aircraft imagery and dense coincident surface observations},
  author={Beck, Richard and Xu, Min and Zhan, Shengan and Liu, Hongxing and Johansen, Richard A and Tong, Susanna and Yang, Bo and Shu, Song and Wu, Qiusheng and Wang, Shujie and others},
  journal={Remote Sensing},
  volume={9},
  number={6},
  pages={538},
  year={2017},
  publisher={MDPI}
}

@article{appel2024efficient,
  title={Efficient data-driven gap filling of satellite image time series using deep neural networks with partial convolutions},
  author={Appel, Marius},
  journal={Artificial Intelligence for the Earth Systems},
  volume={3},
  number={2},
  pages={220055},
  year={2024},
  publisher={American Meteorological Society}
}

@article{appel2020spatiotemporal,
  title={Spatiotemporal multi-resolution approximations for analyzing global environmental data},
  author={Appel, Marius and Pebesma, Edzer},
  journal={Spatial Statistics},
  volume={38},
  pages={100465},
  year={2020},
  publisher={Elsevier}
}

@article{archana2024deep,
  title={Deep learning models for digital image processing: a review},
  author={Archana, R and Jeevaraj, PS Eliahim},
  journal={Artificial Intelligence Review},
  volume={57},
  number={1},
  pages={11},
  year={2024},
  publisher={Springer}
}

@article{barzegar2020short,
  title={Short-term water quality variable prediction using a hybrid CNN--LSTM deep learning model},
  author={Barzegar, Rahim and Aalami, Mohammad Taghi and Adamowski, Jan},
  journal={Stochastic Environmental Research and Risk Assessment},
  volume={34},
  number={2},
  pages={415--433},
  year={2020},
  publisher={Springer}
}

@article{chen2020deep,
  title={A deep learning CNN architecture applied in smart near-infrared analysis of water pollution for agricultural irrigation resources},
  author={Chen, Huazhou and Chen, An and Xu, Lili and Xie, Hai and Qiao, Hanli and Lin, Qinyong and Cai, Ken},
  journal={Agricultural Water Management},
  volume={240},
  pages={106303},
  year={2020},
  publisher={Elsevier}
}

@article{chen2011simple,
  title={A simple and effective method for filling gaps in Landsat ETM+ SLC-off images},
  author={Chen, Jin and Zhu, Xiaolin and Vogelmann, James E and Gao, Feng and Jin, Suming},
  journal={Remote sensing of environment},
  volume={115},
  number={4},
  pages={1053--1064},
  year={2011},
  publisher={Elsevier}
}

@inproceedings{cresson2019optical,
  title={Optical image gap filling using deep convolutional autoencoder from optical and radar images},
  author={Cresson, R{\'e}mi and Ienco, Dino and Gaetano, Raffaele and Ose, Kenji and Minh, D Ho Tong},
  booktitle={IGARSS 2019-2019 IEEE International Geoscience and Remote Sensing Symposium},
  pages={218--221},
  year={2019},
  organization={IEEE}
}

@inproceedings{ehret2021automatic,
  title={Automatic Monitoring of Water Level in Small Lakes Using Planetscope},
  author={Ehret, Thibaud and Lajouanie, Simon and Lefran{\c{c}}ois, Victor and De Franchis, Carlo},
  booktitle={2021 IEEE International Geoscience and Remote Sensing Symposium IGARSS},
  pages={3356--3359},
  year={2021},
  organization={IEEE}
}

@article{frazier2021technical,
  title={A technical review of planet smallsat data: Practical considerations for processing and using planetscope imagery},
  author={Frazier, Amy E and Hemingway, Benjamin L},
  journal={Remote Sensing},
  volume={13},
  number={19},
  pages={3930},
  year={2021},
  publisher={MDPI}
}

@article{gerber2018predicting,
  title={Predicting missing values in spatio-temporal remote sensing data},
  author={Gerber, Florian and de Jong, Rogier and Schaepman, Michael E and Schaepman-Strub, Gabriela and Furrer, Reinhard},
  journal={IEEE Transactions on Geoscience and Remote Sensing},
  volume={56},
  number={5},
  pages={2841--2853},
  year={2018},
  publisher={IEEE}
}

@article{mo2022bayesian,
  title={Bayesian convolutional neural networks for predicting the terrestrial water storage anomalies during GRACE and GRACE-FO gap},
  author={Mo, Shaoxing and Zhong, Yulong and Forootan, Ehsan and Mehrnegar, Nooshin and Yin, Xin and Wu, Jichun and Feng, Wei and Shi, Xiaoqing},
  journal={Journal of Hydrology},
  volume={604},
  pages={127244},
  year={2022},
  publisher={Elsevier}
}

@inproceedings{daniels2022filling,
  title={Filling Cloud Gaps in Satellite AOD Retrievals Using an LSTM CNN-Autoencoder Model},
  author={Daniels, Jacob and Bailey, Colleen P and Liang, Lu},
  booktitle={IGARSS 2022-2022 IEEE International Geoscience and Remote Sensing Symposium},
  pages={2758--2761},
  year={2022},
  organization={IEEE}
}

@inproceedings{scaramuzza2005landsat,
  title={Landsat 7 scan line corrector-off gap-filled product development},
  author={Scaramuzza, Pasquale and Barsi, Julia},
  booktitle={Proceeding of Pecora},
  volume={16},
  number={3},
  pages={23--27},
  year={2005}
}

@article{liang2023reconstructing,
  title={Reconstructing aerosol optical depth using spatiotemporal Long Short-Term Memory convolutional autoencoder},
  author={Liang, Lu and Daniels, Jacob and Biancardi, Michael and Zhou, Yuye},
  journal={Scientific Data},
  volume={10},
  number={1},
  pages={842},
  year={2023},
  publisher={Nature Publishing Group UK London}
}

@inproceedings{liu2018image,
  title={Image inpainting for irregular holes using partial convolutions},
  author={Liu, Guilin and Reda, Fitsum A and Shih, Kevin J and Wang, Ting-Chun and Tao, Andrew and Catanzaro, Bryan},
  booktitle={Proceedings of the European conference on computer vision (ECCV)},
  pages={85--100},
  year={2018}
}

@article{lops2021application,
  title={Application of a partial convolutional neural network for estimating geostationary aerosol optical depth data},
  author={Lops, Yannic and Pouyaei, Arman and Choi, Yunsoo and Jung, Jia and Salman, Ahmed Khan and Sayeed, Alqamah},
  journal={Geophysical Research Letters},
  volume={48},
  number={15},
  pages={e2021GL093096},
  year={2021},
  publisher={Wiley Online Library}
}

@article{mansaray2021comparing,
  title={Comparing PlanetScope to Landsat-8 and Sentinel-2 for sensing water quality in reservoirs in agricultural watersheds},
  author={Mansaray, Abubakarr S and Dzialowski, Andrew R and Martin, Meghan E and Wagner, Kevin L and Gholizadeh, Hamed and Stoodley, Scott H},
  journal={Remote Sensing},
  volume={13},
  number={9},
  pages={1847},
  year={2021},
  publisher={MDPI}
}

@article{moreno2020multispectral,
  title={Multispectral high resolution sensor fusion for smoothing and gap-filling in the cloud},
  author={Moreno-Mart{\'\i}nez, {\'A}lvaro and Izquierdo-Verdiguier, Emma and Maneta, Marco P and Camps-Valls, Gustau and Robinson, Nathaniel and Mu{\~n}oz-Mar{\'\i}, Jordi and Sedano, Fernando and Clinton, Nicholas and Running, Steven W},
  journal={Remote Sensing of Environment},
  volume={247},
  pages={111901},
  year={2020},
  publisher={Elsevier}
}

@article{paheding2024advancing,
  title={Advancing horizons in remote sensing: a comprehensive survey of deep learning models and applications in image classification and beyond},
  author={Paheding, Sidike and Saleem, Ashraf and Siddiqui, Mohammad Faridul Haque and Rawashdeh, Nathir and Essa, Almabrok and Reyes, Abel A},
  journal={Neural Computing and Applications},
  volume={36},
  number={27},
  pages={16727--16767},
  year={2024},
  publisher={Springer}
}

@techreport{Planet2023TechReport,
  author      = {Planet Team},
  title       = {PlanetScope Product Specifications},
  year        = {2023},
  url         = {https://assets.planet.com/docs/Planet_PSScene_Imagery_Product_Spec_letter_screen.pdf}
}

@inproceedings{pritt2017satellite,
  title={Satellite image classification with deep learning},
  author={Pritt, Mark and Chern, Gary},
  booktitle={2017 IEEE applied imagery pattern recognition workshop (AIPR)},
  pages={1--7},
  year={2017},
  organization={IEEE}
}

@article{qian2024gap,
  title={A gap filling method for daily evapotranspiration of global flux data sets based on deep learning},
  author={Qian, Long and Wu, Lifeng and Zhang, Zhitao and Fan, Junliang and Yu, Xingjiao and Liu, Xiaogang and Yang, Qiliang and Cui, Yaokui},
  journal={Journal of Hydrology},
  volume={641},
  pages={131787},
  year={2024},
  publisher={Elsevier}
}

@article{roy2008multi,
  title={Multi-temporal MODIS--Landsat data fusion for relative radiometric normalization, gap filling, and prediction of Landsat data},
  author={Roy, David P and Ju, Junchang and Lewis, Philip and Schaaf, Crystal and Gao, Feng and Hansen, Matt and Lindquist, Erik},
  journal={Remote Sensing of Environment},
  volume={112},
  number={6},
  pages={3112--3130},
  year={2008},
  publisher={Elsevier}
}

@article{sagan2021field,
  title={Field-scale crop yield prediction using multi-temporal WorldView-3 and PlanetScope satellite data and deep learning},
  author={Sagan, Vasit and Maimaitijiang, Maitiniyazi and Bhadra, Sourav and Maimaitiyiming, Matthew and Brown, Davis R and Sidike, Paheding and Fritschi, Felix B},
  journal={ISPRS journal of photogrammetry and remote sensing},
  volume={174},
  pages={265--281},
  year={2021},
  publisher={Elsevier}
}

@article{segal2020cloud,
  title={Cloud detection algorithm for multi-modal satellite imagery using convolutional neural-networks (CNN)},
  author={Segal-Rozenhaimer, Michal and Li, Alan and Das, Kamalika and Chirayath, Ved},
  journal={Remote Sensing of Environment},
  volume={237},
  pages={111446},
  year={2020},
  publisher={Elsevier}
}

@article{shen2015missing,
  title={Missing information reconstruction of remote sensing data: A technical review},
  author={Shen, Huanfeng and Li, Xinghua and Cheng, Qing and Zeng, Chao and Yang, Gang and Li, Huifang and Zhang, Liangpei},
  journal={IEEE Geoscience and Remote Sensing Magazine},
  volume={3},
  number={3},
  pages={61--85},
  year={2015},
  publisher={IEEE}
}

@inproceedings{szegedy2017inception,
  title={Inception-v4, inception-resnet and the impact of residual connections on learning},
  author={Szegedy, Christian and Ioffe, Sergey and Vanhoucke, Vincent and Alemi, Alexander},
  booktitle={Proceedings of the AAAI conference on artificial intelligence},
  volume={31},
  number={1},
  year={2017}
}

@article{tu2022radiometric,
  title={The radiometric accuracy of the 8-band multi-spectral surface reflectance from the planet SuperDove constellation},
  author={Tu, Yu-Hsuan and Johansen, Kasper and Aragon, Bruno and El Hajj, Marcel M and McCabe, Matthew F},
  journal={International Journal of Applied Earth Observation and Geoinformation},
  volume={114},
  pages={103035},
  year={2022},
  publisher={Elsevier}
}

@article{vizzari2022planetscope,
  title={PlanetScope, Sentinel-2, and Sentinel-1 data integration for object-based land cover classification in Google Earth Engine},
  author={Vizzari, Marco},
  journal={Remote Sensing},
  volume={14},
  number={11},
  pages={2628},
  year={2022},
  publisher={MDPI}
}

@article{waldner2020deep,
  title={Deep learning on edge: Extracting field boundaries from satellite images with a convolutional neural network},
  author={Waldner, Fran{\c{c}}ois and Diakogiannis, Foivos I},
  journal={Remote sensing of environment},
  volume={245},
  pages={111741},
  year={2020},
  publisher={Elsevier}
}

@article{wang2012three,
  title={A three-dimensional gap filling method for large geophysical datasets: Application to global satellite soil moisture observations},
  author={Wang, Guojie and Garcia, Damien and Liu, Yi and De Jeu, Richard and Dolman, A Johannes},
  journal={Environmental Modelling \& Software},
  volume={30},
  pages={139--142},
  year={2012},
  publisher={Elsevier}
}

@article{wang2022new,
  title={A new object-class based gap-filling method for PlanetScope satellite image time series},
  author={Wang, Jing and Lee, Calvin KF and Zhu, Xiaolin and Cao, Ruyin and Gu, Yating and Wu, Shengbiao and Wu, Jin},
  journal={Remote Sensing of Environment},
  volume={280},
  pages={113136},
  year={2022},
  publisher={Elsevier}
}

@techreport{WaterNSW2015TechReport,
  author      = {WaterNSW},
  title       = {Fact sheet: Warragamba Dam},
  year        = {2015},
  url         = {https://www.waternsw.com.au/__data/assets/pdf_file/0011/114869/Fact-sheet-Warragamba-Dam-fact-sheet-Nov15-ScreenRes.pdf}
}

@article{xing2022spatiotemporal,
  title={Spatiotemporal reconstruction of MODIS normalized difference snow index products using U-Net with partial convolutions},
  author={Xing, De and Hou, Jinliang and Huang, Chunlin and Zhang, Weimin},
  journal={Remote Sensing},
  volume={14},
  number={8},
  pages={1795},
  year={2022},
  publisher={MDPI}
}

@article{yi2021filling,
  title={Filling the data gaps within GRACE missions using singular spectrum analysis},
  author={Yi, Shuang and Sneeuw, Nico},
  journal={Journal of Geophysical Research: Solid Earth},
  volume={126},
  number={5},
  pages={e2020JB021227},
  year={2021},
  publisher={Wiley Online Library}
}

@article{yin2016gap,
  title={Gap-filling of landsat 7 imagery using the direct sampling method},
  author={Yin, Gaohong and Mariethoz, Gregoire and McCabe, Matthew F},
  journal={Remote Sensing},
  volume={9},
  number={1},
  pages={12},
  year={2016},
  publisher={MDPI}
}

@article{zhang2021rich,
  title={Rich CNN features for water-body segmentation from very high resolution aerial and satellite imagery},
  author={Zhang, Zhili and Lu, Meng and Ji, Shunping and Yu, Huafen and Nie, Chenhui},
  journal={Remote Sensing},
  volume={13},
  number={10},
  pages={1912},
  year={2021},
  publisher={MDPI}
}

@article{golshan2020patterns,
  title={Patterns of cyanobacterial abundance in a major drinking water reservoir: what 3 years of comprehensive monitoring data reveals?},
  author={Golshan, Azadeh and Evans, Craig and Geary, Phillip and Morrow, Abigail and Maeder, Marcel and Tauler, Rom{\`a}},
  journal={Environmental monitoring and assessment},
  volume={192},
  pages={1--11},
  year={2020},
  publisher={Springer}
}

@phdthesis{mueller2014role,
  title={The role of nutrients in cyanobacterial blooms in a shallow reservoir},
  author={Mueller, Stefanie},
  year={2014}
}

@article{water2011catchment,
  title={Catchment Management Plan: Hunter Water’s Eight Element Plan for Our Catchments},
  author={Water, Hunter},
  year={2011},
  publisher={Hunter Water Corporation, Newcastle, Australia}
}

@article{king2022murray,
  title={Murray--Darling Basin 2023 Outlook. Environmental Values: Technical Literature Review},
  author={King, Alison J and Mynott, Julia H and Bond, Nick and Grieger, Rebekah and Hamilton, David and Hawke, Tahneal and Johnston-Bates, Jaiden and Kennard, Mark and King, Georgia and Kingsford, Richard T and others},
  year={2022},
  publisher={La Trobe University, Centre for Freshwater Ecosystems}
}

@article{liu2021effectiveness,
  title={The effectiveness of global constructed shallow waterbody design guidelines to limit harmful algal blooms},
  author={Liu, S and Johnson, F and Tamburic, B and Crosbie, ND and Glamore, W},
  journal={Water Resources Research},
  volume={57},
  number={8},
  pages={e2020WR028918},
  year={2021},
  publisher={Wiley Online Library}
}

@article{vilhena2010role,
  title={The role of climate change in the occurrence of algal blooms: Lake Burragorang, Australia},
  author={Vilhena, Leticia C and Hillmer, Ingrid and Imberger, J{\"o} rg},
  journal={Limnology and Oceanography},
  volume={55},
  number={3},
  pages={1188--1200},
  year={2010},
  publisher={Wiley Online Library}
}

@article{alzubaidi2021review,
  title={Review of deep learning: concepts, CNN architectures, challenges, applications, future directions},
  author={Alzubaidi, Laith and Zhang, Jinglan and Humaidi, Amjad J and Al-Dujaili, Ayad and Duan, Ye and Al-Shamma, Omran and Santamar{\'\i}a, Jos{\'e} and Fadhel, Mohammed A and Al-Amidie, Muthana and Farhan, Laith},
  journal={Journal of big Data},
  volume={8},
  pages={1--74},
  year={2021},
  publisher={Springer}
}

@article{han2023survey,
  title={A survey of machine learning and deep learning in remote sensing of geological environment: Challenges, advances, and opportunities},
  author={Han, Wei and Zhang, Xiaohan and Wang, Yi and Wang, Lizhe and Huang, Xiaohui and Li, Jun and Wang, Sheng and Chen, Weitao and Li, Xianju and Feng, Ruyi and others},
  journal={ISPRS Journal of Photogrammetry and Remote Sensing},
  volume={202},
  pages={87--113},
  year={2023},
  publisher={Elsevier}
}

@article{shirmard2022review,
  title={A review of machine learning in processing remote sensing data for mineral exploration},
  author={Shirmard, Hojat and Farahbakhsh, Ehsan and M{\"u}ller, R Dietmar and Chandra, Rohitash},
  journal={Remote Sensing of Environment},
  volume={268},
  pages={112750},
  year={2022},
  publisher={Elsevier}
}

@article{li2023comprehensive,
  title={A comprehensive survey on design and application of autoencoder in deep learning},
  author={Li, Pengzhi and Pei, Yan and Li, Jianqiang},
  journal={Applied Soft Computing},
  volume={138},
  pages={110176},
  year={2023},
  publisher={Elsevier}
}

@article{boulila2021novel,
  title={A novel CNN-LSTM-based approach to predict urban expansion},
  author={Boulila, Wadii and Ghandorh, Hamza and Khan, Mehshan Ahmed and Ahmed, Fawad and Ahmad, Jawad},
  journal={Ecological Informatics},
  volume={64},
  pages={101325},
  year={2021},
  publisher={Elsevier}
}

@article{pan2023cnn,
  title={A CNN--LSTM machine-learning method for estimating particulate organic carbon from remote sensing in lakes},
  author={Pan, Banglong and Yu, Hanming and Cheng, Hongwei and Du, Shuhua and Cai, Shutong and Zhao, Minle and Du, Juan and Xie, Fazhi},
  journal={Sustainability},
  volume={15},
  number={17},
  pages={13043},
  year={2023},
  publisher={MDPI}
}

@article{shakeel2022aladdin,
  title={ALADDIn: Autoencoder-LSTM-based anomaly detector of deformation in InSAR},
  author={Shakeel, Anza and Walters, Richard J and Ebmeier, Susanna K and Al Moubayed, Noura},
  journal={IEEE Transactions on Geoscience and Remote Sensing},
  volume={60},
  pages={1--12},
  year={2022},
  publisher={IEEE}
}

@article{brezonik2005landsat,
  title={Landsat-based remote sensing of lake water quality characteristics, including chlorophyll and colored dissolved organic matter (CDOM)},
  author={Brezonik, Patrick and Menken, Kevin D and Bauer, Marvin},
  journal={Lake and Reservoir Management},
  volume={21},
  number={4},
  pages={373--382},
  year={2005},
  publisher={Taylor \& Francis}
}

@article{mishra2012normalized,
  title={Normalized difference chlorophyll index: A novel model for remote estimation of chlorophyll-a concentration in turbid productive waters},
  author={Mishra, Sachidananda and Mishra, Deepak R},
  journal={Remote Sensing of Environment},
  volume={117},
  pages={394--406},
  year={2012},
  publisher={Elsevier}
}

@article{ho2017using,
  title={Using Landsat to extend the historical record of lacustrine phytoplankton blooms: a Lake Erie case study},
  author={Ho, Jeff C and Stumpf, Richard P and Bridgeman, Thomas B and Michalak, Anna M},
  journal={Remote Sensing of Environment},
  volume={191},
  pages={273--285},
  year={2017},
  publisher={Elsevier}
}

@article{kingma2014adam,
  title={Adam: A method for stochastic optimization},
  author={Kingma, Diederik P and Ba, Jimmy},
  journal={arXiv preprint arXiv:1412.6980},
  year={2014}
}

@article{alotaibi2020hybrid,
  title={A hybrid deep ResNet and inception model for hyperspectral image classification},
  author={Alotaibi, Bandar and Alotaibi, Munif},
  journal={PFG--Journal of Photogrammetry, Remote Sensing and Geoinformation Science},
  volume={88},
  number={6},
  pages={463--476},
  year={2020},
  publisher={Springer}
}

@article{iyer2021deep,
  title={Deep learning ensemble method for classification of satellite hyperspectral images},
  author={Iyer, Praveen and Sriram, A and Lal, Shyam},
  journal={Remote Sensing Applications: Society and Environment},
  volume={23},
  pages={100580},
  year={2021},
  publisher={Elsevier}
}

@inproceedings{donahue2015long,
  title={Long-term recurrent convolutional networks for visual recognition and description},
  author={Donahue, Jeffrey and Anne Hendricks, Lisa and Guadarrama, Sergio and Rohrbach, Marcus and Venugopalan, Subhashini and Saenko, Kate and Darrell, Trevor},
  booktitle={Proceedings of the IEEE conference on computer vision and pattern recognition},
  pages={2625--2634},
  year={2015}
}

@inproceedings{vinyals2015show,
  title={Show and tell: A neural image caption generator},
  author={Vinyals, Oriol and Toshev, Alexander and Bengio, Samy and Erhan, Dumitru},
  booktitle={Proceedings of the IEEE conference on computer vision and pattern recognition},
  pages={3156--3164},
  year={2015}
}

@inproceedings{srivastava2015unsupervised,
  title={Unsupervised learning of video representations using lstms},
  author={Srivastava, Nitish and Mansimov, Elman and Salakhudinov, Ruslan},
  booktitle={International conference on machine learning},
  pages={843--852},
  year={2015},
  organization={PMLR}
}

@inproceedings{abadi2016tensorflow,
  title={$\{$TensorFlow$\}$: a system for $\{$Large-Scale$\}$ machine learning},
  author={Abadi, Mart{\'\i}n and Barham, Paul and Chen, Jianmin and Chen, Zhifeng and Davis, Andy and Dean, Jeffrey and Devin, Matthieu and Ghemawat, Sanjay and Irving, Geoffrey and Isard, Michael and others},
  booktitle={12th USENIX symposium on operating systems design and implementation (OSDI 16)},
  pages={265--283},
  year={2016}
}

@article{dekker2002analytical,
  title={Analytical algorithms for lake water TSM estimation for retrospective analyses of TM and SPOT sensor data},
  author={Dekker, Arnold G and Vos, RJ and Peters, SWM},
  journal={International journal of remote sensing},
  volume={23},
  number={1},
  pages={15--35},
  year={2002},
  publisher={Taylor \& Francis}
}

@article{wang2006applications,
  title={Applications of Landsat-5 TM imagery in assessing and mapping water quality in Reelfoot Lake, Tennessee},
  author={Wang, F and Han, L and Kung, H-T and Van Arsdale, RB},
  journal={International Journal of Remote Sensing},
  volume={27},
  number={23},
  pages={5269--5283},
  year={2006},
  publisher={Taylor \& Francis}
}

@inproceedings{schroeder2008remote,
  title={Remote sensing of apparent and inherent optical properties of Tasmanian coastal waters: application to MODIS data},
  author={Schroeder, Th and Brando, VE and Cherukuru, N and Clementson, L and Blondeau-Patissier, D and Dekker, AG and Schaale, M and Fischer, J},
  booktitle={Proceedings of the XIX Ocean Optics Conference, Barga, Italy},
  pages={6--10},
  year={2008}
}

@article{gao1996ndwi,
  title={NDWI—A normalized difference water index for remote sensing of vegetation liquid water from space},
  author={Gao, Bo-Cai},
  journal={Remote sensing of environment},
  volume={58},
  number={3},
  pages={257--266},
  year={1996},
  publisher={Elsevier}
}

@article{mcfeeters1996use,
  title={The use of the Normalized Difference Water Index (NDWI) in the delineation of open water features},
  author={McFeeters, Stuart K},
  journal={International journal of remote sensing},
  volume={17},
  number={7},
  pages={1425--1432},
  year={1996},
  publisher={Taylor \& Francis}
}

@article{chen2022remote,
  title={Remote sensing big data for water environment monitoring: Current status, challenges, and future prospects},
  author={Chen, Jinyue and Chen, Shuisen and Fu, Rao and Li, Dan and Jiang, Hao and Wang, Chongyang and Peng, Yongshi and Jia, Kai and Hicks, Brendan J},
  journal={Earth's Future},
  volume={10},
  number={2},
  pages={e2021EF002289},
  year={2022},
  publisher={Wiley Online Library}
}

@article{sagan2020monitoring,
  title={Monitoring inland water quality using remote sensing: Potential and limitations of spectral indices, bio-optical simulations, machine learning, and cloud computing},
  author={Sagan, Vasit and Peterson, Kyle T and Maimaitijiang, Maitiniyazi and Sidike, Paheding and Sloan, John and Greeling, Benjamin A and Maalouf, Samar and Adams, Craig},
  journal={Earth-Science Reviews},
  volume={205},
  pages={103187},
  year={2020},
  publisher={Elsevier}
}

@article{caballero2021monitoring,
  title={Monitoring cyanoHABs and water quality in Laguna Lake (Philippines) with Sentinel-2 satellites during the 2020 Pacific typhoon season},
  author={Caballero, Isabel and Navarro, Gabriel},
  journal={Science of the Total Environment},
  volume={788},
  pages={147700},
  year={2021},
  publisher={Elsevier}
}

@article{zhi2024heterogeneity,
  title={Heterogeneity and influencing factors of algal blooms in the reservoir-impacted tributary: Evidence from remote sensing and physical-based model},
  author={Zhi, Xiaosha and Chen, Lei and Chen, Shibo and Yu, Jiaqi and Jiang, Jing and Xu, Yanzhe and Li, Leifang and Meng, Xinyi and Shen, Zhenyao},
  journal={Journal of Hydrology},
  volume={634},
  pages={131058},
  year={2024},
  publisher={Elsevier}
}

@article{aubriot2020assessing,
  title={Assessing the origin of a massive cyanobacterial bloom in the R{\'\i}o de la Plata (2019): Towards an early warning system},
  author={Aubriot, Luis and Zabaleta, Bernardo and Bordet, Facundo and Sienra, Daniel and Risso, Jimena and Achkar, Marcel and Somma, Andrea},
  journal={Water Research},
  volume={181},
  pages={115944},
  year={2020},
  publisher={Elsevier}
}

@article{chen2022novel,
  title={A novel big data mining framework for reconstructing large-scale daily MAIAC AOD data across China from 2000 to 2020},
  author={Chen, Binjie and Ye, Yang and Tong, Cheng and Deng, Jinsong and Wang, Ke and Hong, Yang},
  journal={GIScience \& Remote Sensing},
  volume={59},
  number={1},
  pages={670--685},
  year={2022},
  publisher={Taylor \& Francis}
}





\end{document}